\documentclass[11pt]{article}

\usepackage[final]{acl}
\usepackage{enumitem}
\usepackage{times}
\usepackage{latexsym}
\usepackage{multirow}
\usepackage{algorithmic}
\usepackage{algorithm}
\usepackage{marvosym}

\usepackage[T1]{fontenc}
\usepackage{microtype}

\usepackage{inconsolata}

\usepackage{graphicx}

\title{APPSI-139: A Parallel Corpus of English Application Privacy Policy Summarization and Interpretation}

\author{
Pengyun~Zhu$^{1,2}$, 
Qiheng~Sun$^{2}$, 
Long~Wen$^{2}$, Yanbo~Wang$^{3}$, 
Yang Cao$^{4}$,
\\\textbf{Junxu~Liu}$^{5}$,
\textbf{Deyi~Xiong}$^{1}$,
\textbf{Jinfei~Liu}$^{2,6}$\thanks{Corresponding Author.}, \textbf{Zhibo~Wang}$^{2}$, \textbf{Kui~Ren}$^{2}$ \\
$^{1}$ Tianjin University, $^{2}$ Zhejiang University, 
$^{3}$ North University of China,\\
$^{4}$ Institute of Science Tokyo,
$^{5}$ The Hong Kong Polytechnic University,
\\
$^{6}$ Hangzhou High-Tech Zone (Binjiang) Institute of Blockchain and Data Security\\
\texttt{\{pengyunzhu, dyxiong\}@tju.edu.cn}, 
\texttt{jinfeiliu@zju.edu.cn\textsuperscript{\Letter}} \\
}

\begin{document}
\maketitle
\begin{abstract}
Privacy policies are essential for users to understand how service providers handle their personal data. However, these documents are often long and complex, as well as filled with technobabble and legalese, causing users to unknowingly accept terms that may even contradict the law. While summarizing and interpreting these privacy policies is crucial, there is a lack of high-quality English parallel corpus optimized for legal clarity and readability.
To address this issue, we introduce APPSI-139, a high-quality English privacy policy corpus meticulously annotated by domain experts, specifically designed for summarization and interpretation tasks. The corpus includes 139 English privacy policies, 15,692 rewritten parallel corpora, and 36,351 fine-grained annotation labels across 11 data practice categories. Concurrently, we propose TCSI-pp-V2, a hybrid privacy policy  summarization and interpretation framework that employs an alternating training strategy and coordinates multiple expert modules to effectively balance computational efficiency and accuracy. Experimental results show that the hybrid summarization system built on APPSI-139 corpus and the TCSI-pp-V2 framework outperform large language models, such as GPT-4o and LLaMA-3-70B, in terms of readability and reliability. The source code and dataset are available at \url{https://github.com/EnlightenedAI/APPSI-139}.
\end{abstract}
\section{Introduction}

The privacy policy outlines how service providers collect, process, store, manage, and use the personal information of individuals interacting with their applications~\citep{NEURIPS2023_capp130}. Service providers are permitted to handle personal information according to the stipulations outlined in the agreement, with the authorization of users.

However, most privacy policies are often considered ``incomprehensible'' due to complex technical jargon, legal language, and convoluted grammatical structures~\citep{ermakova2014privacy, singh2011user}. This issue is further exacerbated by ``rational ignorance,'' where users perceive the effort required to understand lengthy privacy policies as disproportionate to the benefits, and ``dark patterns,'' where design elements subtly guide users to quickly click ``Agree'' or ``Join Now'' without fully understanding the consequences~\citep{zhuhou2018}. Such behavior can lead to the disclosure or misuse of users' sensitive data without explicit consent~\citep{obar2020biggest}. Although efforts such as ``Privacy Nutrition Labels''~\citep{DBLP:conf/chi/LiRACH22,DBLP:journals/corr/abs-2204-03556}, ``LPL''~\citep{gerl2018lpl}, and ``TILT''~\citep{grunewald2021tilt} aim to establish standardized and formalized guidelines for privacy policies, assisting service providers in presenting their privacy policies more clearly and understandably, ensuring that service providers genuinely adhere to these guidelines remains a significant challenge.

Automatic summarization is considered a potential solution, utilizing natural language processing and machine learning techniques to automatically extract and summarize key information from text~\citep{keymanesh2022adapative,zayed2025automatic,fu-etal-2023-inverse}. However, due to the lack of parallel corpora for interpreting privacy policies, most existing methods for summarizing privacy policies focus primarily on extracting key information and shortening the length of the policy, while giving less attention to addressing the ``incomprehensibility'' caused by legal and technical terminology.

In response to the above issues, we have organized domain experts to carefully annotate a parallel corpus for English privacy policy summarization and interpretation, APPSI-139, and based on this, we propose the privacy policy summarization and interpretation framework, Tsci-pp-V2.

The APPSI-139 corpus includes English privacy policies from 139 top applications across domains such as shopping, live streaming, sports, and gaming. It has been annotated by legal experts to ensure the content is comprehensive, accurate, and easy to understand. The corpus consists of two main components: high-quality annotations and a policy interpretation parallel corpus. Specifically, there are 36,351 annotations covering 11 categories of data practices and 3 special tags, used to identify core clauses, label sensitive information, and recognize potential risks. Additionally, the policy interpretation parallel corpus contains 15,692 pairs, where legal experts have rewritten and paraphrased key clauses containing technobabble and legalese into comprehensible language to reduce the difficulty of understanding.

The TCSI-pp-V2 framework is a Topic-Controlled framework for Summarizing and Interpreting Privacy Policy, built on end-to-end multi-task learning. The framework decomposes the summarization task into five subtasks and employs an alternating training strategy across five parameter-sharing experts, each specialized in one subtask. 

Specifically, four experts are dedicated to identifying and locating key clauses by detecting their Importance, Risk, Sensitivity, and Topic, while a Rewrite expert is responsible for rewriting and interpreting technobabble and legalese. Through this collaborative architecture, the framework enables precise clause identification and transformation into comprehensible language, supporting users in making informed and cognitively sound privacy decisions. 

For clarity, we summarize our main contributions as follows.

\begin{itemize}[leftmargin=*]
\item[$\bullet$]{We created APPSI-139, the first parallel corpus of English privacy policies, meticulously annotated and explained by legal experts, aimed at providing user-friendly interpretations to help users make informed privacy decisions.}
\item[$\bullet$]{We propose the TCSI-pp-V2 framework, which coordinates multiple expert modules through an alternating training strategy, effectively balancing computational efficiency with accuracy.}
\item[$\bullet$]{Based on APPSI-139 corpus and the TCSI-pp-V2 framework, we develop a summarization and interpretation system. Questionnaire survey evaluations show that, compared to general-purpose models like GPT-4o, our system outperforms in terms of readability and reliability, helping users make more informed and rational privacy decisions.}
\end{itemize}

\begin{figure*}[t]\centering 
    \includegraphics[width=0.88\textwidth,trim={0cm 0cm 0cm 0cm}, clip]{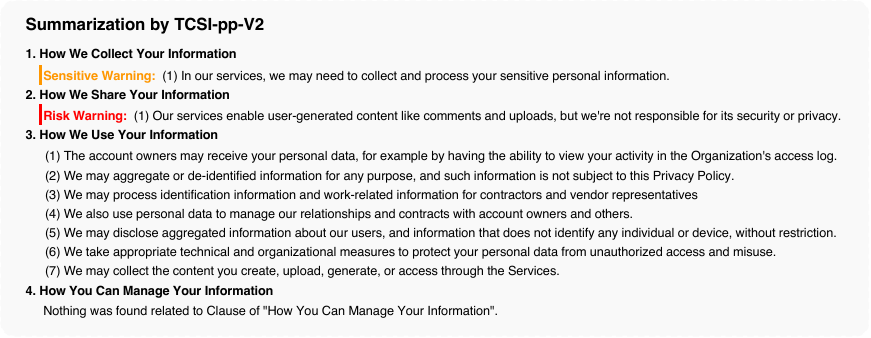}
    \caption{Summarization by TCSI-pp-V2.}
    \label{fig:our}
    \vspace{-0em}
\end{figure*}

\section{Related Work}\label{shown}
\subsection{Privacy Policy Corpus}

In the current research landscape, we observe that publicly available corpus for privacy policies are not only limited in number but also predominantly focused on information extraction or content analysis. For instance, OPP-115~\citep{wilson2016creation} and APP-350~\citep{zimmeck2019maps}  are among the earliest known privacy policy corpora, which were compiled using expert annotation and crowd-sourced annotation respectively. These corpora break down privacy policies into finer data practices, providing useful resources for tasks such as classification and content extraction of privacy policy texts.
PI-Extract~\citep{bui2021automated} is a special project that selects 30 privacy policies from OPP-115 and annotates a more detailed text range for entity recognition tasks. Furthermore, PrivacyQA~\citep{ravichander2019question} and PolicyQA~\citep{ahmad2020policyqa} are privacy policy question-and-answer corpus based on OPP-130. These two corpus consist of pairs made up of the ``original text'' of the privacy policy and ``questions'', which can extract relevant sentences or paragraphs based on the user's questions.
Optoutchoice-2020~\citep{bannihatti2020finding} and Optoutchoice-2017~\citep{sathyendra2017identifying} are two corpora specifically annotated with content related to privacy settings, such as ad tracking, which play an important role in helping users extract and manage privacy settings in privacy policies.
\citet{huang2024analyzing} use large language models (LLMs) to create a large-scale classification corpus.
The Chinese corpus CA4P-483~\citep{zhao2022fine} was annotated by trained students and includes sentence-level category labels and detailed data processing practices. These privacy policy corpora are mainly designed for extracting specific information or content. CAPP-130~\citep{NEURIPS2023_capp130} introduced a summary corpus aimed at addressing the ``incomprehensibility” issue through rewriting, with detailed sentence-level annotations and rewrite examples. However, the English version of CAPP-130 was obtained through machine translation, without adaptation to native English expressions, and lacks the linguistic precision and legal clarity required for effective understanding of privacy policies.

\subsection{Privacy Policy Summarization}
Recently, natural language processing and machine learning technologies have made significant strides in addressing the readability issues of ``length'' and ``incomprehensibility'' in privacy policies. These technologies offer an efficient solution by automatically extracting key information or generating condensed summaries. Specifically, there are three main approaches: extractive summarization, abstractive summarization, and hybrid summarization.
\textit{Extractive summarization} improves the readability of privacy policies by selecting sentences or paragraphs related to specific topics. For example, \citet{wilson2016creation,harkous2018polisis,liu2018towards,tomuro2016automatic,zhao2022fine,ravichander2019question,ahmad2020policyqa} and PolicyGPT \citep{tang2023policygpt} tackle policy verbosity by extracting topic-relevant sentences.
Meanwhile,  \citet{keymanesh2020toward,nokhbeh2022privacycheck,nokhbeh2020privacycheck} focuses on using machine learning to assess and retrieve content that may pose potential privacy risks. 
\textit{Abstractive summarization} leverages general text simplification techniques \citep{al2021automated, alva2020data,siddharthan2014survey} to generate concise and paraphrased summaries.
Recent advances in large language models (LLMs)\citep{saini2025text,DBLP:journals/corr/abs-2411-11072,DBLP:journals/corr/abs-2407-03884}, such as ChatGPT\citep{ouyang2022training}, open new opportunities for privacy policy analysis by simplifying complex linguistic structures \citep{feng2023sentence}. However, fine-tuned smaller models for specific tasks provide more practical advantages\citep{qin2023chatgpt}, while LLMs remain constrained by computational resources, deployment costs, and the high demands of prompt design when handling lengthy texts \citep{zhang2025systematic,DBLP:journals/inffus/FangPYMCTL26,DBLP:journals/corr/abs-2412-17686} .
\textit{Hybrid summarization} combine the strengths of extractive and abstractive approaches, improving the controllability and consistency of generative models \citep{wang2017integrating}. For example, the TCSI-pp framework \citep{NEURIPS2023_capp130} outperforms GPT-4 in addressing privacy policy ``length'' and ``incomprehensibility'' but still faces computational redundancy due to independent encoding.

\section{The APPSI-139 Corpus and TCSI-pp-V2 Framework}\label{sec:preliminaries}
In this section, we disclose the details of APPSI-139 in Section \ref{subsec: dataset} and present the summarization framework TCSI-pp-V2 in Section \ref{subsec:model}. The summarization examples are shown in Figure \ref{fig:our}.

\subsection{The APPSI-139 Corpus}\label{subsec: dataset}

\begin{figure}[t]\centering 
    \includegraphics[width=0.42\textwidth]{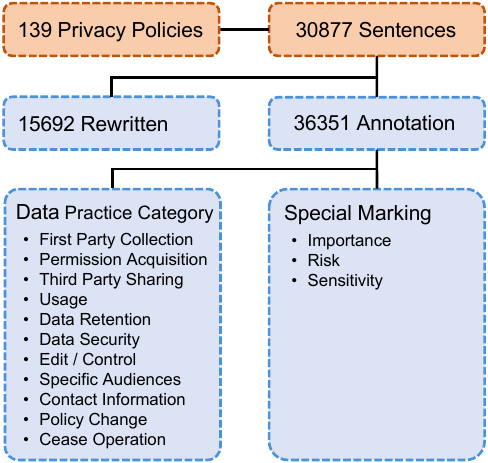}
    \caption{The organization of APPSI-139.}
    \label{fig:capp}
    \vspace{-0.5em}
\end{figure}

Privacy policies often contain a large amount of technical jargon, which requires a strong professional background to fully comprehend. To ensure that the annotations accurately capture the core elements of privacy policies, a team of legal experts—including a law professor and five lawyers with master’s degrees—has developed annotation guidelines for English privacy policy summaries through multiple rounds of pilot annotations. Building on the design of CAPP-130~\citep{NEURIPS2023_capp130}, the guidelines are divided into three parts: \textit{Data Practice Categories}, \textit{Special Markings}, and \textit{Rewritten Sentences}. Figure~\ref{fig:capp} illustrates the organization of APPSI-139.

\textbf{\emph{Data Practice Categories}} classify and organize information within privacy policies, ensuring clearer and more standardized comprehension of data handling practices. These categories include First Party Collection, Permission Acquisition, Third Party Sharing, Usage, Data Retention, Data Security, Edit/Control, Specific Audiences, Contact Information, Policy Change, and Cease Operation. Detailed definitions of each category are provided in Appendix \ref{Data_Practice}.

\textbf{\emph{Special Marking}} highlights clauses related to important data practices, sensitive personal information, and potential risks, including \emph{Importance}, \emph{Risk}, and \emph{Sensitivity} tags. The definitions are as follows:
\begin{itemize}[leftmargin=*]
\item[$\bullet$]{\textbf{Importance}: Applies to provisions directly related to key data processing elements, including the definitions of Risk,'' Sensitivity,'' and the eleven categories of ``Data Practices'' in the clauses. These provisions remain crucial for subsequent summarization and rewriting tasks.}
\item[$\bullet$]{\textbf{Risk}: Applies to content that may raise compliance issues or violate major data protection regulations (e.g., GDPR, CCPA), including vague language about data handling, such as ``We may share your location data with third parties without further notice''.}
\item[$\bullet$]{\textbf{Sensitivity}: Applies to data practices involving users' sensitive personal information, including biometric identifiers, precise geolocation, and financial account details. The sensitivity annotation aligns with several standards, including the EU General Data Protection Regulation (GDPR), the US National Institute of Standards and Technology (NIST) SP 800-122 guidelines, and the Chinese standard GB/T 35273-2020, each representing regional and national regulations. For details, see Appendix \ref{psi}.}
\end{itemize}

\textbf{\emph{Rewritten Sentence}} provides user-friendly and comprehensible versions by rewriting and paraphrasing sentences labeled as Importance'', Sensitive'', and ``Risk''.

To ensure that privacy policies reflect the latest trends and legal requirements, while also maintaining representativeness and diversity, we source the privacy policies of the top 100 most downloaded applications from two major app marketplaces: \emph{Google Play} and the \emph{App Store} in the English-speaking world. After excluding duplicates and instances where different applications share the same privacy policy, we obtain 139 distinct and representative privacy policies. These documents, current as of October 2023, cover a wide range of mainstream application types, including shopping, live streaming, sports, navigation, and various game genres such as real-time battles, sports competitions, simulation management, and board games. Given that subtle semantic differences in privacy policies can have legal implications, and that smaller models are still widely used in edge deployments, we employ regular expressions to segment these privacy policies into individual sentences for detailed annotation and interpretation.

Finally, we permit each segmented sentence to have multiple labels. The annotation work for this English privacy policy dataset is carried out by five experts, each holding a master's degree with a legal background and possessing lawyer certification. They possess advanced English language skills, with proficiency surpassing the TOEFL and IELTS graduate admission requirements. Among these five annotators, three have previously participated in the Chinese privacy policy annotation of the CAPP-130 dataset\citep{NEURIPS2023_capp130}. Before starting this project, all annotators undergo systematic training covering annotation guidelines and typical examples to ensure annotation quality and stability. 
A preliminary analysis of five annotated documents yields a \textit{Cohen's Kappa} score of 0.892, indicating strong agreement. To account for the multi-label and multi-task nature of the dataset, this score was calculated by considering all classification categories, including Special Marking, Importance, Risk, and Sensitivity. We computed the agreement across these tasks to ensure the reliability of the integrated schema. This high level of agreement informs our decision to adopt the strategy of assigning each expert to a single subtask, which includes the five specific tasks: identifying sentences related to Importance'', Risk'', Sensitivity'', Data Practice Categories'', or those requiring a ``Rewritten''. During formal annotation, to transparently handle edge cases, annotators document ambiguous clauses. These are then resolved through discussion, with final adjudication by senior reviewers.

This results in a high-quality corpus named APPSI-139 (\underline{A}pplication \underline{P}rivacy \underline{P}olicies \underline{S}ummarization and \underline{I}nterpretation, set of 139). This corpus comprises 36,351 annotations for 30,877 sentences, including 15,692 rewritten sentences.

\begin{table}[t]
\centering

\begin{tabular}{ccccc}
\hline
\small
{\color{black}\textbf{Topic}}
 & \textbf{Num} & \textbf{Pct} & \textbf{Med.} & \textbf{Avg.} \\ 
\hline
First  & 3667 & 11.8\% & 23.0 & 27.2\\
Permission  & 153 & 0.5\% & 22.0 & 23.2\\
Third & 3019 & 9.8\% & 25.0 & 29.0\\
Usage & 3159 & 10.2\% & 23.0 & 28.8 \\
Retention & 974 & 3.2\% & 25.0 & 28.2\\
Security & 1069 & 3.5\% & 23.0 & 25.4\\
Specific & 1858 & 6.0\% &23.0 & 25.9\\
Control & 3772 & 12.2\% & 22.0 & 24.8\\
Contact & 1229 & 4.0\% &19.0 & 21.0\\
Change & 795 & 2.7\% & 18.0 & 20.2\\
Cease & 13 & 0.04\% & 23.0 & 31.2\\
Important & 15795 & 51.04\% &24.0 & 27.3\\
Risk & 577 & 1.9\% & 26.0 & 29.7\\ 
Sensitivity &374 &1.21\% & 26.0 & 28.9\\
\hline

\end{tabular}
\caption{Statistical Information of APPSI-139. }
\label{tab:dataset-statistics}
\end{table}

\begin{figure}[t]\centering 
\includegraphics[width=0.45\textwidth]{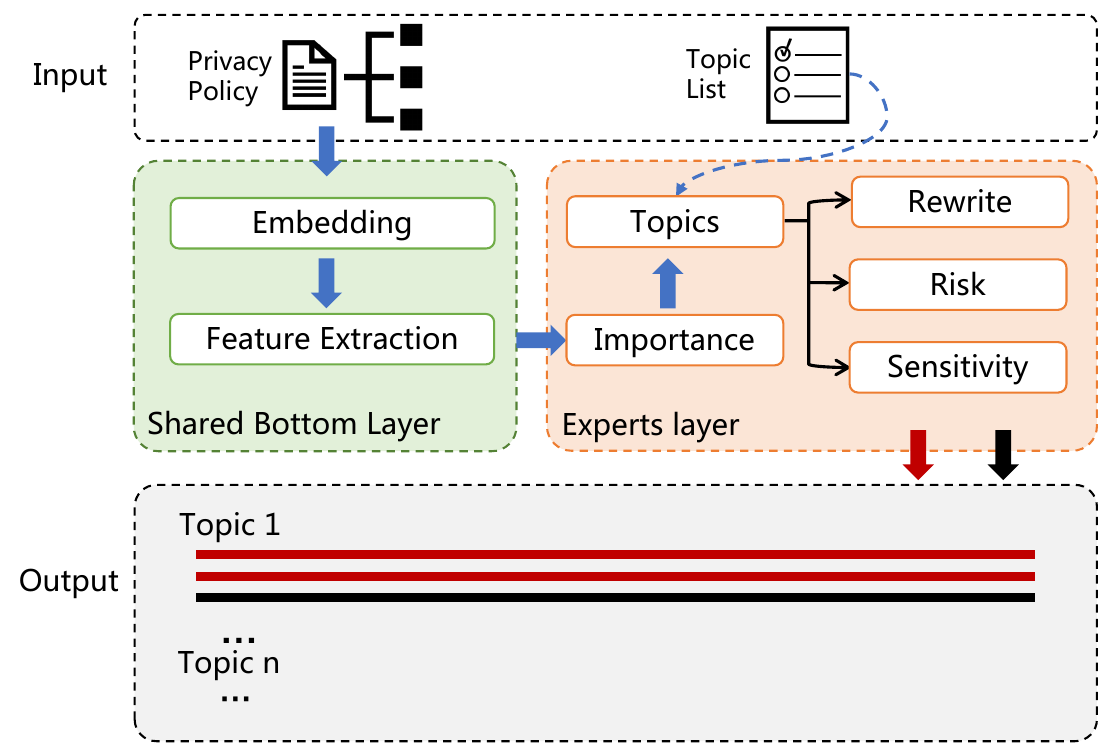}\vspace{-0em}
    \caption{The framework of TCSI-pp-V2.}
    \label{fig:process}
\end{figure}

The corpus enables analyze the composition of typical English privacy policies from a data practice perspective. Table~\ref{tab:dataset-statistics} presents the statistics for APPSI-139. The Pct (\emph{percentage}) represents the proportion of sentences in a specific data practice category or with special markers relative to the total number of sentences. Since a sentence can have multiple labels, the total percentage may exceed 100\%. \emph{Mean} and \emph{median} represent the average and middle values of sentence lengths, respectively. The average sentence length exceeds the median for all categories, indicating a right-skewed distribution. The APPSI-139 corpus shows an uneven distribution of data practice categories in English privacy policies. First-party collection (11.8\%), third-party sharing (9.8\%), editing/control (12.2\%), and usage (10.2\%) account for about 44\% of the content, forming the core components. Rewritten sentences average 20 words, compared to 27 in the originals, reflecting a 26\% reduction. In APPSI-139, the ``Risk'' label appears in 1.9\% of sentences, highlighting the widespread privacy and data security risks faced by users of English-language applications.

\subsection{The TCSI-pp-V2 Framework}\label{subsec:model}

To ensure the completeness and controllability of privacy policy summarization, we propose TCSI-pp-V2 (\underline{T}opic-\underline{C}ontrolled framework for \underline{S}ummarization and \underline{I}nterpretation of \underline{P}rivacy \underline{P}olicies), which represents a hybrid summarization approach based on end-to-end multi-task learning and builds upon the APPSI-139 corpus. As shown in Figure \ref{fig:process}, the framework consists of four primary components:

\textbf{\emph{Input}}:
First, we preprocess any given English privacy policy $\forall P$ using \emph{regular expressions} to extract a collection of sentences at the individual sentence level, denoted as $P = \{p_1, \ldots, p_n\}$, where $p_n$ refers to the $n$-th sentence in the policy. Second, users select the \emph{topics} they wish to summarize from the \emph{topics} that we provide.

\textbf{\emph{Shared Bottom Layer}}: The Shared Bottom Layer processes the input signals and extracts key features for all tasks, reducing redundancy by learning shared representations. It begins with an embedding layer that captures the semantic content of the input text, represented as $P = \{p_1, \ldots, p_n\}$. This layer converts the text into a numerical matrix, denoted as $E = \{e_1, \ldots, e_n\}$. Next, the feature extraction function, $F_f(e_j, \theta_f)$, processes the embeddings $E$ and extracts features, producing a new set, $Features = \{f_1, \ldots, f_n\}$.

\textbf{\emph{Experts Layer}}: Expert layers are components trained for specific tasks, with each expert layer specifically designed to handle different aspects of privacy policy summarization and interpretation. Specifically, we train five distinct expert layers to accomplish the task of privacy policy summarization.
\begin{itemize}[leftmargin=*]
\item[$\bullet$]{\textbf{Importance identification} $F_i(f_j, \theta_i)$ takes all $\forall p$ corresponding features $f_j$ as input and identifies the ``Important'' sentences. Here, $\theta_i$ represents a set of parameters obtained through training.}
\item[$\bullet$]{\textbf{Topics identification} $F_t(f_j, \theta_t)$ indexes the features $f_j$ of $\forall p_j | important = true$ to determine the data practice category (topic) to which these ``important'' sentences belong. Based on user-specified $topics$, we extract sentences consistent with them, producing the set $Filtered$.}
\item[$\bullet$]{\textbf{Risk identification} $F_r(f_j, \theta_r)$ indexes the features $f_j$ of $\forall p_j | p \in Filtered $ to identify sentences that potentially harm user data security.}
\item[$\bullet$]{\textbf{Sensitivity identification} $F_s(f_j, \theta_s)$ indexes the features $f_j$ of  $\forall p_j | p \in Filtered$ to flag sentences related to sensitive personal information.}
\item[$\bullet$]{\textbf{Text interpretation} $F_{rewrite}(f_j, \theta_{rewrite})$ rewrites important long sentences containing obscure professional and technical terms into simple and easily understandable short sentences. This addresses the issue of ``incomprehensibility''. Specifically, $F_{rewrite}(f_j, \theta_{rewrite})$ uses an auto-regressive model $P(z_t | f_j; z_{1:t-1})$ to rephrase the original privacy policy into formulations that are easily comprehensible for the general public.}
\end{itemize}

\textbf{\emph{Output}}: The framework summarizes and interprets privacy policies based on preselected topics and highlights sentences that contain Risk'' and Sensitivity'' information to enhance their readability.

In summary, TCSI-pp-V2 introduces a shared encoder across all sub-tasks, replacing the five separate encoders in TCSI-pp, and incorporates task-specific logic optimization, thereby theoretically reducing encoding overhead by about 80\%. Pseudocode appears in Appendix~\ref{pseudocode}.

\begin{table*}[t]\centering

{
\begin{tabular}{cccccccccc}
\hline
{\color{black}\multirow{2}{*}{TCSI-pp}} &{\color{black}\multirow{2}{*}{ Method}} &\multicolumn{2}{c}{Topics}& \multicolumn{2}{c}{Important} & \multicolumn{2}{c}{Risk} & \multicolumn{2}{c}{Sensitive}\\

&{\color{black}} & Micro &	Macro   & Micro &	Macro & Micro &	Macro& Micro&	Macro\\
\hline
\multirow{10}{*}{V1}&RoBERTa	&80.77	&79.09	&69.95    &69.90    &95.27	 &61.60 &97.28 &71.53\\

&Bert	&81.00	 &79.63	&76.87    &76.21    &95.54	&61.58 &97.07&62.51\\


&alBert		&73.28  &72.34 	& 75.09    &73.94    &97.82	&60.99&97.82&60.98\\
&Pert	&79.82	 &78.58	&78.84   &78.20    &96.74	&49.17 &97.93 &49.48\\

&ERNIE	&81.43	 &79.56	& 78.29  &77.55    &96.41	& 62.41 &97.06 &69.25\\
&DistilBert	&80.55	 &79.18	& 77.73    &77.03    &95.70	& 58.18 &97.34 &60.43\\
&DeBERTa	&78.47	 &76.33	& 79.35    &78.34    &99.27	& 66.55 &97.01 &68.33\\
&XlNet &81.39	 &80.13	& 77.90    &77.36    & 96.03    &59.73 &96.90 &66.45\\
&Electra &81.91	 &80.03	& 78.00    &77.22    &96.08	& 60.69 &96.90 &62.20\\
&\textbf{Mean}&\textbf{79.85}&\textbf{78.31}&\textbf{76.89}&\textbf{76.19}&\textbf{96.54}&\textbf{60.01}&\textbf{97.26}&\textbf{63.46} \\ 
\hline
\multirow{5}{*}{V2}&mT5&78.18	 &77.12	&73.93     &73.55    &	95.60& 58.77&96.96 &67.40 \\
&XlNet2gpt&80.79	 &78.79	& 76.25    &74.99    &95.98	&63.39 &96.68 &64.09 \\
&Electra2gpt&80.00	 &78.83	& 76.19    &75.79    &95.49	&59.31 &93.36 &64.53 \\
&Bert2gpt&79.68	 &77.97	& 76.74    &75.81    &95.81	& 60.80 &96.74 &63.45 \\

&\textbf{Mean} &\textbf{79.66}&\textbf{78.17}&\textbf{75.78}&\textbf{75.04}&\textbf{95.72}&\textbf{60.57}&\textbf{95.93}&\textbf{64.86}\\
\hline
\end{tabular}}
\caption{{\color{black}Evaluation metrics for the information identification models.}}\label{tab:information extraction}
\label{tab:fine screening} 
\end{table*}%

\begin{table*}[t]\centering

\begin{tabular}{c c c c c c c c}
\hline
{\color{black} TCSI-pp}&{\color{black} Model} & ROUGE-1 &	ROUGE-2  &	ROUGE-L & BERT-score & Bart-score \\
\hline
\multirow{5}{*}{V1}&mT5	&\textbf{0.7199}	&\textbf{0.6296}	&\textbf{0.7072} &\textbf{0.947} &\textbf{-1.56} \\
&Bert2gpt	&0.6477	&0.5411	&0.6312   &0.930 &-2.22 \\
&Bert2bert	&0.6784	&0.5980	&0.6610   &0.939 &-2.14\\
&ERNIE2gpt	&0.5805	&0.3931	&0.5459   &0.926 &-2.26\\
&\textbf{Mean}&\textbf{0.6566}&\textbf{0.5405}&\textbf{0.6363} &\textbf{0.935}&\textbf{-2.05}\\
\hline
\multirow{5}{*}{V2}&mT5 &\textbf{0.7051}	&\textbf{0.6111}	&\textbf{0.6903} &\textbf{0.943} &\textbf{-1.68}\\
&Bert2gpt &0.6548&0.5456&0.6375&0.933&-2.03\\ 
&XlNet2gpt &0.6751&0.5623&0.6598 &0.939&-1.786\\
&Electra2gpt &0.6653 & 0.5476& 0.6491&0.938    &-1.89 \\
&\textbf{Mean}&\textbf{0.6752}&\textbf{0.5666}&\textbf{0.6592}&\textbf{0.938}&\textbf{-1.84} \\ 
\hline
\end{tabular}
\caption{Evaluation metrics for the rewrite expert models.}\label{tab:Interpretations}
\end{table*}%

\section{Experiments} \label{Experiments}

In this section, we introduced the experimental setup in Section \ref{subsec:Baseline Methods}, presented performance benchmarks for information extraction models based on APPSI-139 in Section \ref{subsec:Information Extraction}, discussed performance benchmarks for the rewriting model in Section \ref{subsec:Interpretations}, assessed TCSI-pp-V2's readability through a social survey in Section \ref{subsec:Readability}, conducted a comprehensive evaluation of resource efficiency in Section \ref{subsec:Resource Efficiency}, performed a comparative analysis with state-of-the-art Large Language Models in Section \ref{subsec:llm_comparison}, and investigated the model's robustness to varying input lengths in Section \ref{subsec:length_analysis}.

\subsection{Base Models}\label{subsec:Baseline Methods}

We have implemented the TCSI-pp-V2 framework using mT5, Bert2GPT, XLNet2GPT, and Electra2GPT as base models. The model architecture has consisted of an encoder followed by four classification heads for the respective sub-tasks and a decoder for content rewriting. This design has allowed for the joint optimization of classification and generation tasks within a unified framework.
For comparison, we also have implemented the TCSI-pp framework. For the four information identification sub-tasks, we have trained models based on the following Transformer encoders: RoBERTa (\emph{xlm-roberta-base})\citep{DBLP:journals/corr/abs-1911-02116}, BERT (\emph{bert-base-cased})\citep{DBLP:journals/corr/abs-1810-04805}, ERNIE (\emph{ernie-2.0-base-en})\citep{sun2019ernie20}, PERT (\emph{english-pert-large})\citep{cui2022pert}, DistilBERT (\emph{distilbert-base-uncased})\citep{Sanh2019DistilBERTAD}, ELECTRA (\emph{electra-base-discriminator})\citep{clark2020electra}, XLNet (\emph{xlnet-base-cased})\citep{DBLP:journals/corr/abs-1906-08237}, DeBERTa (\emph{deberta-v3-base})\citep{he2021debertav3}, ALBERT (\emph{albert-base-v2})\citep{DBLP:journals/corr/abs-1909-11942}, and mT5 (\emph{mT5-small})\citep{xue2020mt5}. For the content rewriting sub-task, we have trained the TCSI-pp framework using Bert2Bert, Bert2GPT, ERNIE2GPT, and mT5. All of these base models have been available from the Hugging Face model repository HuggingFace\footnote{\url{https://huggingface.co/}}.
To ensure that the input sequence does not exceed the maximum position limit of the pre-trained models, we have truncated documents by removing the ending part. Data splitting details appear in Appendix \ref{Splitting}.

\subsection{Information Extraction}\label{subsec:Information Extraction}
Table \ref{tab:information extraction} has presented the performance of models trained on the APPSI-139 corpus for information extraction. The table has included four multi-task learning TCSI-pp-V2 models and nine single-task learning TCSI-pp baseline models. The tasks Importance,'' Risk,'' and Sensitivity'' have been binary classification, while Topic'' has involved multi-class classification. Due to data sparsity, the categories Cease Operation'' and Permission Acquisition'' have been excluded. Evaluation metrics have included Micro-F1 and Macro-F1 scores.

As shown in Table~\ref{tab:information extraction}, under the TCSI-pp framework, DeBERTa performs best on binary classification, while XLNet and ELECTRA excel in multi-class tasks. In TCSI-pp-V2, XLNet2GPT slightly outperforms the other three models. Across single-task models (BERT, XLNet, DeBERTa, ELECTRA) and their multi-task counterparts, performance differences remain within 0.02, indicating that TCSI-pp-V2's multi-task models match the performance of TCSI-pp while reducing computational costs. Further details regarding Precision, Recall, and F1 scores for the multi-class ``Topic'' classification are provided in Appendix \ref{app:topic_extraction}.

\begin{table*}[t]\centering

{
\begin{tabular}{c c c c c c c c c c c c}
\hline
{\color{black} Model} & Q1 &	Q2  &Q3 & Q4 & Q5 &Q6& Q7 &Q8  &Q9 & Q10 &Percentage\\

\hline

TCSI-pp-V2	&\textbf{22} &\textbf{27} &\textbf{27}  &20 &\textbf{20} &\textbf{20} &14 &14 &\textbf{20} &\textbf{23} &\textbf{39.06\%}\\
Llama3-70b	&18	&6	&13  &5  &15 &11 &\textbf{21} &\textbf{21} &14 &10 &25.28\%\\
GPT-4o	    &5	&15	&6   &\textbf{25} &12 &11 &13 &13 &14 &16 &24.52\%\\
Kimi	    &8	&5	&7   &3  &6  &11 &5  &5  &5  &4  &11.13\%\\



\hline
\end{tabular}}
\caption{{\color{black} Questionnaire statistics of readability.}}\label{tab:Readability}
\vspace{-0.5em}
\end{table*}%

\subsection{Rewritten and Interpretations}\label{subsec:Interpretations}

To evaluate the performance of the interpretations model in TCSI-pp-V2, we have used rewritten sentences from APPSI-139 for fine-tuning. Four interpretations models have been implemented based on the TCSI-pp and TCSI-pp-V2 frameworks, respectively.
Table \ref{tab:Interpretations} has displayed the evaluation metrics for information identification models, including \emph{ROUGE}~\citep{lin-2004-ROUGE}, \emph{Bert-score}~\citep{zhang2019bertscore}, and \emph{Bart-score}~\citep{NEURIPS2021_e4d2b6e6}. These metrics provide a multi-dimensional assessment: ROUGE measures lexical overlap, while Bert-score and Bart-score capture semantic consistency and generative quality. Together, they robustly reflect the model's alignment with human expert judgments and legal intent beyond simple word-matching.

Table~\ref{tab:Interpretations} shows that mT5 outperforms other models in both the single-task setting based on TCSI-pp and the multi-task setting based on TCSI-pp-V2, with a slight advantage in single-task training. However, for Bert2GPT, the model trained with multi-task learning surpasses the single-task version across all metrics. Moreover, in the rewriting task, multi-task models consistently achieve higher average scores than single-task models, with particularly notable improvements in accuracy and recall. This suggests that multi-task training maintains or even enhances performance in generation tasks, as related tasks share common features and the model benefits from learning more generalized representations through joint training.

\subsection{Readability Questionnaire Survey of TCSI-pp-V2}\label{subsec:Readability}

To evaluate the readability of the privacy policy summarization system based on TCSI-pp-V2, we have conducted a social survey on WJX\footnotemark[2]\footnotetext[2]{\url{https://www.wjx.cn/}}. The survey has consisted of 10 single-choice questions, each presenting a paragraph from a privacy policy alongside summaries generated by four models: our model, CPT-4o\footnotemark[3]\footnotetext[3]{\url{https://openai.com/index/hello-gpt-4o/}}, Llama3-70b\footnotemark[4]\footnotetext[4]{\url{https://huggingface.co/meta-llama/Meta-Llama-3-70B}}, and Kimi (Moonshot V1)\footnotemark[5]\footnotetext[5]{\url{https://kimi.moonshot.cn/}}. We have invited participants to select the summary they have found most readable, based on criteria such as comprehensibility, completeness, fidelity, and conciseness. In total, we have collected 53 valid responses, as shown in Table~\ref{tab:Readability}. The respondents, aged between 18 and 40, all have received university-level education. The summarization examples, instruction templates, and detailed statistics of the survey participants are provided in Appendix~\ref{ars}.

According to Table \ref{tab:Readability}, 39.06\% of participants rate TCSI-pp-V2 summaries as the most readable based on comprehensibility, completeness, fidelity, and conciseness, followed by Llama3-70b (25.28\%), GPT-4o (24.52\%), and Kimi (11.13\%). Additionally, TCSI-pp-V2 leads in seven out of ten questions, outperforming the other models overall. These results indicate that TCSI-pp-V2 demonstrates excellent overall readability when summarizing application privacy policies, outperforming Llama3-70b, GPT-4o, and other LLMs in privacy policy summarization and interpretation task.

\subsection{Resource Efficiency}\label{subsec:Resource Efficiency}

To evaluate the computational efficiency and practical deployability of our framework, we have measured the GPU memory usage and inference latency of TCSI-pp and TCSI-pp-V2 (both utilizing the mT5-small backbone). The evaluation has been conducted on a representative subset of 10 privacy policies comprising 1,893 sentences, with the metrics detailed in Table \ref{tab:efficiency}.

\begin{table}[t]
\centering
\small
\begin{tabular}{ccc}
\hline
\textbf{Metric} & \textbf{TCSI-pp} & \textbf{TCSI-pp-V2} \\ \hline
Encoding Time (s) & 92.66 & 24.72 \\
Task Inference Time (s)& 99.28 & 98.54\\
Total Time (s) & 191.94 & 123.26 \\
Avg. Time per Sentence (s) & 0.101 & 0.065 \\
Inference Memory (MB) & 7343 & 2766 \\ \hline
\end{tabular}
\caption{GPU Memory and Time Cost Comparison.}
\label{tab:efficiency}
\end{table}

The results demonstrate that TCSI-pp-V2 achieves substantial efficiency gains. Specifically, the encoding time is reduced by approximately 73\%, and the average processing time per sentence is nearly halved. Furthermore, the peak GPU memory overhead significantly drops from 7,343 MB to 2,766 MB. These improvements confirm that our optimized shared-encoder architecture effectively mitigates computational redundancy, making it more suitable for real-time applications without sacrificing inference accuracy.

\begin{table*}[ht]
\centering
\begin{tabular}{llccccc}
\hline
\textbf{Task} &\textbf{Metric} & \textbf{Qwen3-8B} & \textbf{Llama3-8B} & \textbf{GPT-4o-mini} & \textbf{Gemini-2.5} & \textbf{TCSI-pp-V2} \\  \hline
\multirow{2}{*}{Topic}&Micro F1     & 47.85 & 30.77 & 50.36 & 65.38 & \textbf{78.18} \\
&Macro F1     & 44.44 & 24.65 & 32.83 & 6.42  & \textbf{77.12} \\
\multirow{2}{*}{Important}&Micro F1 & 63.50 & 52.53 & 51.00 & 73.33 & \textbf{73.93} \\
&Macro F1 & 62.11 & 52.45 & 49.36 & 71.29 & \textbf{73.55} \\
\multirow{2}{*}{Risk}&Micro F1      & 85.53 & \textbf{96.00} & 89.34 & 85.05 & 95.60 \\
&Macro F1      & 49.30 & 48.98 & 51.24 & 45.95 & \textbf{58.77} \\
\multirow{2}{*}{Sensitive}&Micro F1 & 45.37 & 28.01 & 38.64 & 23.33 & \textbf{96.96} \\
&Macro F1 & 32.52 & 21.88 & 27.87 & 21.76 & \textbf{67.40} \\
\multirow{3}{*}{Rewritten}&ROUGE-L            & 0.4541 & 0.4156 & 0.4286 & 0.4776 & \textbf{0.6903} \\
&BERTScore          & 0.8970 & 0.8520 & 0.8950 & 0.9070 & \textbf{0.9430} \\
&BARTScore          & -2.76  & -3.03  & -2.89  & -2.78  & \textbf{-1.68}  \\ \hline
\end{tabular}
\caption{Performance Comparison between LLMs and TCSI-pp-V2 on the APPSI-139 Dataset.}
\label{tab:llm_comparison}
\end{table*}

\subsection{Comparative Analysis with Large Language Models}\label{subsec:llm_comparison}

To benchmark our framework against contemporary state-of-the-art baselines, we have compared TCSI-pp-V2 with several prominent Large Language Models (LLMs), including open-weight models (Qwen3-8B, Llama-3.1-8B) and proprietary systems (GPT-4o-mini, Gemini-2.5). These models have been evaluated on the APPSI-139 dataset using meticulously designed prompt engineering to handle both classification and abstractive rewriting tasks, with results shown in Table \ref{tab:llm_comparison}.

As shown in Table \ref{tab:llm_comparison}, TCSI-pp-V2 consistently outperforms the selected LLMs across most tasks and does so with a significantly smaller parameter count. This indicates that even though general-purpose LLMs possess vast general knowledge, they still struggle with the domain-specific nuances of privacy legalities, particularly in Topic and Sensitive classification. Moreover, our model maintains superior generation quality with higher ROUGE-L and BERTScore values. These results validate that specialized fine-tuning on expert-annotated data remains more effective than zero-shot prompting for high-precision legal analysis.

\subsection{Robustness to Input Length}\label{subsec:length_analysis}

To assess the robustness of TCSI-pp-V2, we have conducted a comprehensive analysis focusing on extreme input distributions. Specifically, TCSI-pp-V2 has been evaluated on the 100 longest and shortest test samples across all classification and rewriting tasks. This evaluation has highlighted the framework's adaptability to varying information densities, with detailed results in Table \ref{tab:length_analysis}.

The experimental results demonstrate that TCSI-pp-V2 maintains high stability across the input length spectrum. We observe only minor performance fluctuations: exceptionally long policies exhibit a slight decrease in Risk classification accuracy due to their complex clause structures, while extremely short texts yield marginally lower ROUGE-L scores due to the scarcity of contextual cues. Nevertheless, the consistent performance across all subsets underscores the effectiveness of our multi-task architecture in processing diverse and structurally complex privacy documents.

\begin{table}[t]
\centering 
\small
\begin{tabular}{lcccc} 
\hline
Task & Metric & Longest & Shortest & All \\ \hline
\multirow{2}{*}{Topic}&Micro F1     & 79.41 & 76.23 & 78.18 \\
&Macro F1     & 77.20 & 75.09 & 77.12 \\
\multirow{2}{*}{Important}&Micro F1 & 74.56 & 73.64 & 73.93 \\
&Macro F1 & 71.85 & 72.35 & 73.55 \\
\multirow{2}{*}{Risk}&Micro F1      & 94.74 & 94.34 & 95.60 \\
&Macro F1      & 59.39 & 56.91 & 58.77 \\
\multirow{2}{*}{Sensitive} &Micro F1 & 96.83 & 96.18 & 96.96 \\
&Macro F1 & 66.92 & 66.76 & 67.40 \\
\multirow{3}{*}{Rewritten}&ROUGE-L            & 0.6979 & 0.6542 & 0.6903 \\
&BERTScore          & 0.9450 & 0.9430 & 0.9430 \\
&BARTScore          & -1.63  & -1.75  & -1.68  \\ \hline
\end{tabular}
\caption{Performance of Different Lengths.}
\label{tab:length_analysis}
\vspace{-0.5em}
\end{table}

\section{Conclusion}\label{sec:conclusion}

To address the challenges of understanding complex and legally dense privacy policies, we introduce APPSI-139, the first parallel corpus of English privacy policies annotated by legal experts, aimed at providing user-friendly interpretations. We also propose the TCSI-pp-V2 framework, a hybrid summarization method based on multi-task learning that effectively balances computational efficiency and accuracy. Based on this framework, we develop a privacy policy summarization system, which, according to evaluations, outperforms general-purpose models like GPT-4o in readability and reliability. This system helps users make more informed and rational privacy decisions, enhancing the comprehension and interpretation of English privacy policies.

\section*{Limitation}
Currently limited to English-language privacy policies, we aim to expand our framework to cover additional jurisdictions, multilingual settings and other privacy document types in the future. Additionally, as the system relies on model inference and training data, it may, in some cases, overlook key information or generate different interpretations, which could impact the completeness and accuracy of the summary. This is a common risk associated with all generative models when summarizing privacy  policies.

\section*{Acknowledgement}
The present research was supported by the National Key Research and Development Program of China (Grant No. 2023YFE0116400), NSFC (U23A20306, 62472378, U25A20430), the Zhejiang Provincial Natural Science Foundation for Distinguished Young Scholars (LR25F020001), and the Zhejiang Province Pioneer Plan (2024C01074, 2025C01084). We would like to thank the anonymous reviewers for their insightful comments.
\bibliography{ref}

@inproceedings{NEURIPS2023_capp130,
 author = {Zhu, pengyun and Wen, Long and Liu, Jinfei and Xue, Feng and Lou, Jian and Wang, Zhibo and Ren, Kui},
 booktitle = {Advances in Neural Information Processing Systems},
 editor = {A. Oh and T. Naumann and A. Globerson and K. Saenko and M. Hardt and S. Levine},
 pages = {46773--46785},
 publisher = {Curran Associates, Inc.},
 title = {CAPP-130: A Corpus of Chinese Application Privacy Policy Summarization and Interpretation},
 url = {https://proceedings.neurips.cc/paper_files/paper/2023/file/92225ec7e87b97a9e007ca6ab7944b14-Paper-Datasets_and_Benchmarks.pdf},
 volume = {36},
 year = {2023}
}

@article{DBLP:journals/corr/abs-2407-03884,
  author       = {Zhigen Li and
                  Jianxiang Peng and
                  Yanmeng Wang and
                  Tianhao Shen and
                  Minghui Zhang and
                  Linxi Su and
                  Shang Wu and
                  Yihang Wu and
                  Yuqian Wang and
                  Ye Wang and
                  Wei Hu and
                  Jianfeng Li and
                  Shaojun Wang and
                  Jing Xiao and
                  Deyi Xiong},
  title        = {Planning with Large Language Models for Conversational Agents},
  journal      = {CoRR},
  volume       = {abs/2407.03884},
  year         = {2024},
  url          = {https://doi.org/10.48550/arXiv.2407.03884},
  doi          = {10.48550/ARXIV.2407.03884},
  eprinttype   = {arXiv},
  eprint       = {2407.03884},
  bibsource    = {dblp computer science bibliography, https://dblp.org}
}

@article{zayed2025automatic,
  title={Automatic Text Summarization: A Review of Approaches, Challenges, and Future Directions},
  author={Zayed, Sara and Ezzat, Mostafa and Hefny, Hesham A},
  journal={Journal of Computer Science \& Technology},
  volume={25},
  year={2025}
}

@inproceedings{saini2025text,
  title={Text Summarization with LLM: A Comparison of Transformer and Non-Transformer Models},
  author={Saini, Shivanshu and Pathak, Sanskriti and Singh, Suruchi and Bhardwaj, Bharat and Sharma, Saransh},
  booktitle={2025 3rd International Conference on Communication, Security, and Artificial Intelligence (ICCSAI)},
  volume={3},
  pages={23--28},
  year={2025},
  organization={IEEE}
}

@article{zhang2025systematic,
  title={A systematic survey of text summarization: From statistical methods to large language models},
  author={Zhang, Haopeng and Yu, Philip S and Zhang, Jiawei},
  journal={ACM Computing Surveys},
  volume={57},
  number={11},
  pages={1--41},
  year={2025},
  publisher={ACM New York, NY}
}

@article{DBLP:journals/corr/abs-2412-17686,
  author       = {Dan Shi and
                  Tianhao Shen and
                  Yufei Huang and
                  Zhigen Li and
                  Yongqi Leng and
                  Renren Jin and
                  Chuang Liu and
                  Xinwei Wu and
                  Zishan Guo and
                  Linhao Yu and
                  Ling Shi and
                  Bojian Jiang and
                  Deyi Xiong},
  title        = {Large Language Model Safety: {A} Holistic Survey},
  journal      = {CoRR},
  volume       = {abs/2412.17686},
  year         = {2024},
  url          = {https://doi.org/10.48550/arXiv.2412.17686},
  doi          = {10.48550/ARXIV.2412.17686},
  eprinttype   = {arXiv},
  eprint       = {2412.17686},
  bibsource    = {dblp computer science bibliography, https://dblp.org}
}

@article{DBLP:journals/inffus/FangPYMCTL26,
  author       = {Yu Fang and
                  Xiaoqi Pang and
                  Qiyang Yu and
                  Fan Min and
                  Xuemei Cao and
                  Pan Tao and
                  Tianrui Li},
  title        = {Alignment in large vision language models: {A} survey},
  journal      = {Inf. Fusion},
  volume       = {133},
  pages        = {104294},
  year         = {2026},
  url          = {https://doi.org/10.1016/j.inffus.2026.104294},
  doi          = {10.1016/J.INFFUS.2026.104294},
  bibsource    = {dblp computer science bibliography, https://dblp.org}
}

@article{DBLP:journals/corr/abs-2411-11072,
  author       = {Shaolin Zhu and
                  Supryadi and
                  Shaoyang Xu and
                  Haoran Sun and
                  Leiyu Pan and
                  Menglong Cui and
                  Jiangcun Du and
                  Renren Jin and
                  Ant{\'{o}}nio Branco and
                  Deyi Xiong},
  title        = {Multilingual Large Language Models: {A} Systematic Survey},
  journal      = {CoRR},
  volume       = {abs/2411.11072},
  year         = {2024},
  url          = {https://doi.org/10.48550/arXiv.2411.11072},
  doi          = {10.48550/ARXIV.2411.11072},
  eprinttype   = {arXiv},
  eprint       = {2411.11072},
  bibsource    = {dblp computer science bibliography, https://dblp.org}
}

@inproceedings{fu-etal-2023-inverse,
    title = "Inverse Reinforcement Learning for Text Summarization",
    author = "Fu, Yu  and
      Xiong, Deyi  and
      Dong, Yue",
    editor = "Bouamor, Houda  and
      Pino, Juan  and
      Bali, Kalika",
    booktitle = "Findings of the Association for Computational Linguistics: EMNLP 2023",
    month = dec,
    year = "2023",
    address = "Singapore",
    publisher = "Association for Computational Linguistics",
    url = "https://aclanthology.org/2023.findings-emnlp.436/",
    doi = "10.18653/v1/2023.findings-emnlp.436",
    pages = "6559--6570"
}

@inproceedings{sun2019ernie20,
  author       = {Yu Sun and
                  Shuohuan Wang and
                  Yu{-}Kun Li and
                  Shikun Feng and
                  Hao Tian and
                  Hua Wu and
                  Haifeng Wang},
  title        = {{ERNIE} 2.0: {A} Continual Pre-Training Framework for Language Understanding},
  booktitle    = {The Thirty-Fourth {AAAI} Conference on Artificial Intelligence, {AAAI}
                  2020, The Thirty-Second Innovative Applications of Artificial Intelligence
                  Conference, {IAAI} 2020, The Tenth {AAAI} Symposium on Educational
                  Advances in Artificial Intelligence, {EAAI} 2020, New York, NY, USA,
                  February 7-12, 2020},
  pages        = {8968--8975},
  publisher    = {{AAAI} Press},
  year         = {2020},
  url          = {https://doi.org/10.1609/aaai.v34i05.6428},
  doi          = {10.1609/AAAI.V34I05.6428},
  bibsource    = {dblp computer science bibliography, https://dblp.org}
}

@inproceedings{huang2024analyzing,
  title={Analyzing Corporate Privacy Policies using AI Chatbots},
  author={Huang, Ziyuan and Tang, Jiaming and Karir, Manish and Liu, Mingyan and Sarabi, Armin},
  booktitle={Proceedings of the 2024 ACM on Internet Measurement Conference},
  pages={505--515},
  year={2024}
}

@article{tang2023policygpt,
  author       = {Chenhao Tang and
                  Zhengliang Liu and
                  Chong Ma and
                  Zihao Wu and
                  Yiwei Li and
                  Wei Liu and
                  Dajiang Zhu and
                  Quanzheng Li and
                  Xiang Li and
                  Tianming Liu and
                  Lei Fan},
  title        = {PolicyGPT: Automated Analysis of Privacy Policies with Large Language
                  Models},
  journal      = {CoRR},
  volume       = {abs/2309.10238},
  year         = {2023},
  url          = {https://doi.org/10.48550/arXiv.2309.10238},
  doi          = {10.48550/ARXIV.2309.10238},
  eprinttype   = {arXiv},
  eprint       = {2309.10238},
  bibsource    = {dblp computer science bibliography, https://dblp.org}
}

@inproceedings{DBLP:journals/corr/abs-1911-02116,
  author       = {Alexis Conneau and
                  Kartikay Khandelwal and
                  Naman Goyal and
                  Vishrav Chaudhary and
                  Guillaume Wenzek and
                  Francisco Guzm{\'{a}}n and
                  Edouard Grave and
                  Myle Ott and
                  Luke Zettlemoyer and
                  Veselin Stoyanov},
  editor       = {Dan Jurafsky and
                  Joyce Chai and
                  Natalie Schluter and
                  Joel R. Tetreault},
  title        = {Unsupervised Cross-lingual Representation Learning at Scale},
  booktitle    = {Proceedings of the 58th Annual Meeting of the Association for Computational
                  Linguistics, {ACL} 2020, Online, July 5-10, 2020},
  pages        = {8440--8451},
  publisher    = {Association for Computational Linguistics},
  year         = {2020},
  url          = {https://doi.org/10.18653/v1/2020.acl-main.747},
  doi          = {10.18653/V1/2020.ACL-MAIN.747},
  bibsource    = {dblp computer science bibliography, https://dblp.org}
}

@article{DBLP:journals/corr/abs-1810-04805,
  author    = {Jacob Devlin and
               Ming{-}Wei Chang and
               Kenton Lee and
               Kristina Toutanova},
  title     = {{BERT}: Pre-training of Deep Bidirectional Transformers for Language
               Understanding},
  journal   = {CoRR},
  volume    = {abs/1810.04805},
  year      = {2018},
  url       = {http://arxiv.org/abs/1810.04805},
  archivePrefix = {arXiv},
  eprint    = {1810.04805},
  bibsource = {dblp computer science bibliography, https://dblp.org}
}

@inproceedings{he2021debertav3,
  author       = {Pengcheng He and
                  Jianfeng Gao and
                  Weizhu Chen},
  title        = {DeBERTaV3: Improving DeBERTa using ELECTRA-Style Pre-Training with
                  Gradient-Disentangled Embedding Sharing},
  booktitle    = {The Eleventh International Conference on Learning Representations,
                  {ICLR} 2023, Kigali, Rwanda, May 1-5, 2023},
  publisher    = {OpenReview.net},
  year         = {2023},
  url          = {https://openreview.net/forum?id=sE7-XhLxHA},
  bibsource    = {dblp computer science bibliography, https://dblp.org}
}

@inproceedings{clark2020electra,
  author       = {Kevin Clark and
                  Minh{-}Thang Luong and
                  Quoc V. Le and
                  Christopher D. Manning},
  title        = {{ELECTRA}: Pre-training Text Encoders as Discriminators Rather Than
                  Generators},
  booktitle    = {8th International Conference on Learning Representations,
                  Addis Ababa, Ethiopia, April 26-30, 2020},
  url          = {https://openreview.net/forum?id=r1xMH1BtvB},
  bibsource    = {dblp computer science bibliography, https://dblp.org},
    year={2020},
}

@article{Sanh2019DistilBERTAD,
  author       = {Victor Sanh and
                  Lysandre Debut and
                  Julien Chaumond and
                  Thomas Wolf},
  title        = {DistilBERT, a distilled version of {BERT}: Smaller, faster, cheaper
                  and lighter},
  journal      = {CoRR},
  volume       = {abs/1910.01108},
  year         = {2019},
  url          = {http://arxiv.org/abs/1910.01108},
  eprinttype   = {arXiv},
  eprint       = {1910.01108},
  bibsource    = {dblp computer science bibliography, https://dblp.org}
}

@inproceedings{DBLP:journals/corr/abs-1906-08237,
  author       = {Zhilin Yang and
                  Zihang Dai and
                  Yiming Yang and
                  Jaime G. Carbonell and
                  Ruslan Salakhutdinov and
                  Quoc V. Le},
  editor       = {Hanna M. Wallach and
                  Hugo Larochelle and
                  Alina Beygelzimer and
                  Florence d'Alch{\'{e}}{-}Buc and
                  Emily B. Fox and
                  Roman Garnett},
  title        = {{XLNet}: Generalized Autoregressive Pretraining for Language Understanding},
  booktitle    = {Advances in Neural Information Processing Systems 32: Annual Conference
                  on Neural Information Processing Systems 2019, NeurIPS 2019, December
                  8-14, 2019, Vancouver, BC, Canada},
  pages        = {5754--5764},
  year         = {2019},
  url          = {https://proceedings.neurips.cc/paper/2019/hash/dc6a7e655d7e5840e66733e9ee67cc69-Abstract.html},
  bibsource    = {dblp computer science bibliography, https://dblp.org}
}

@article{DBLP:journals/corr/abs-1909-11942,
  author    = {Zhenzhong Lan and
               Mingda Chen and
               Sebastian Goodman and
               Kevin Gimpel and
               Piyush Sharma and
               Radu Soricut},
  title     = {{ALBERT:} {A} Lite {BERT} for Self-supervised Learning of Language
               Representations},
  journal   = {CoRR},
  volume    = {abs/1909.11942},
  year      = {2019},
  url       = {http://arxiv.org/abs/1909.11942},
  archivePrefix = {arXiv},
  eprint    = {1909.11942},
  bibsource = {dblp computer science bibliography, https://dblp.org}
}

@article{obar2020biggest,
  title={The biggest lie on the internet: Ignoring the privacy policies and terms of service policies of social networking services},
  author={Obar, Jonathan A and Oeldorf-Hirsch, Anne},
  journal={Information, Communication \& Society},
  volume={23},
  number={1},
  pages={128--147},
  year={2020},
  publisher={Taylor \& Francis}
}

@inproceedings{keymanesh2020toward,
  title={Toward Domain-Guided Controllable Summarization of Privacy Policies},
  author={Keymanesh, Moniba and Elsner, Micha and Sarthasarathy, Srinivasan},
  booktitle={NLLP@KDD},
  pages={18--24},
  year={2020}
}

@phdthesis{keymanesh2022adapative,
  title={Adapative Summarization for Low-resource Domains and Algorithmic Fairness},
  author={Keymanesh, Moniba},
  year={2022},
  school={The Ohio State University}
}

@article{zhuhou2018,
  title={An Empirical Study on Privacy Policy Reading Intention 
of Social Media Users},
  author={Zhu, Hou and Zhang, Mingxin and Lu, Yonghe },
  journal={Journal of the China Society for Scientific and Technical Information},
  volume={37},
  number={4},
  pages={362--371},
  year={2018}
}

@inproceedings{ermakova2014privacy,
  title={Privacy Policies and Users’ Trust: Does Readability Matter?},
  author={Ermakova, Tatiana and Baumann, Annika and Fabian, Benjamin and Krasnova, Hanna},
  booktitle={Americas Conference on Information Systems},
  year={2014}
}

@article{singh2011user,
  title={A user-centric evaluation of the readability of privacy policies in popular web sites},
  author={Singh, Ravi Inder and Sumeeth, Manasa and Miller, James},
  journal={Information Systems Frontiers},
  volume={13},
  pages={501--514},
  year={2011},
  publisher={Springer}
}

@inproceedings{tomuro2016automatic,
  title={Automatic summarization of privacy policies using ensemble learning},
  author={Tomuro, Noriko and Lytinen, Steven and Hornsburg, Kurt},
  booktitle={Proceedings of the Sixth ACM Conference on Data and Application Security and Privacy},
  pages={133--135},
  year={2016}
}

@inproceedings{ravichander2019question,
  title={Question answering for privacy policies: Combining computational and legal perspectives},
  author={Ravichander, Abhilasha and Black, Alan W and Wilson, Shomir and Norton, Thomas and Sadeh, Norman},
  booktitle={Proceedings of the 2019 Conference on Empirical Methods in Natural Language Processing and the 9th International Joint Conference on Natural Language Processing (EMNLP-IJCNLP)},
  pages={4947--4958},
  year={2019}
}

@inproceedings{ahmad2020policyqa,
  author       = {Wasi Uddin Ahmad and
                  Jianfeng Chi and
                  Yuan Tian and
                  Kai{-}Wei Chang},
  editor       = {Trevor Cohn and
                  Yulan He and
                  Yang Liu},
  title        = {PolicyQA: {A} Reading Comprehension Dataset for Privacy Policies},
  booktitle    = {Findings of the Association for Computational Linguistics: {EMNLP}
                  2020, Online Event, 16-20 November 2020},
  series       = {Findings of {ACL}},
  pages        = {743--749},
  publisher    = {Association for Computational Linguistics},
  year         = {2020},
  url          = {https://doi.org/10.18653/v1/2020.findings-emnlp.66},
  doi          = {10.18653/V1/2020.FINDINGS-EMNLP.66},
  bibsource    = {dblp computer science bibliography, https://dblp.org}
}

@article{feng2023sentence,
  author       = {Yutao Feng and
                  Jipeng Qiang and
                  Yun Li and
                  Yunhao Yuan and
                  Yi Zhu},
  title        = {Sentence Simplification via Large Language Models},
  journal      = {CoRR},
  volume       = {abs/2302.11957},
  year         = {2023},
  url          = {https://doi.org/10.48550/arXiv.2302.11957},
  doi          = {10.48550/ARXIV.2302.11957},
  eprinttype   = {arXiv},
  eprint       = {2302.11957},
  bibsource    = {dblp computer science bibliography, https://dblp.org}
}

@article{siddharthan2014survey,
  title={A survey of research on text simplification},
  author={Siddharthan, Advaith},
  journal={ITL-International Journal of Applied Linguistics},
  volume={165},
  number={2},
  pages={259--298},
  year={2014},
  publisher={John Benjamins}
}

@article{alva2020data,
  title={Data-driven sentence simplification: Survey and benchmark},
  author={Alva-Manchego, Fernando and Scarton, Carolina and Specia, Lucia},
  journal={Computational Linguistics},
  volume={46},
  number={1},
  pages={135--187},
  year={2020}
}

@article{al2021automated,
  title={Automated text simplification: a survey},
  author={Al-Thanyyan, Suha S and Azmi, Aqil M},
  journal={ACM Computing Surveys (CSUR)},
  volume={54},
  number={2},
  pages={1--36},
  year={2021},
  publisher={ACM New York, NY, USA}
}

@article{bui2021automated,
  title={Automated Extraction and Presentation of Data Practices in Privacy Policies},
  author={Bui, Duc and Shin, Kang G and Choi, Jong-Min and Shin, Junbum},
  journal={Proc. Priv. Enhancing Technol.},
  volume={2021},
  number={2},
  pages={88--110},
  year={2021}
}

@inproceedings{wang2017integrating,
  title={Integrating extractive and abstractive models for long text summarization},
  author={Wang, Shuai and Zhao, Xiang and Li, Bo and Ge, Bin and Tang, Daquan},
  booktitle={2017 IEEE International Congress on Big Data (BigData Congress)},
  pages={305--312},
  year={2017},
  organization={IEEE}
}

@misc{doccano,
  title={{doccano}: Text Annotation Tool for Human},
  url={https://github.com/doccano/doccano},
  note={Software available from https://github.com/doccano/doccano},
  author={
    Hiroki, Nakayama and Takahiro, Kubo and Junya, Kamura and Yasufumi, Taniguchi and Xu, Liang},
  year={2018},
}

@inproceedings{qin2023chatgpt,
  title={Is ChatGPT a general-purpose natural language processing task solver?},
  author={Qin, Chengwei and Zhang, Aston and Zhang, Zhuosheng and Chen, Jiaao and Yasunaga, Michihiro and Yang, Diyi},
  booktitle={Proceedings of the 2023 conference on empirical methods in natural language processing},
  pages={1339--1384},
  year={2023}
}

@inproceedings{wilson2016creation,
  title={The creation and analysis of a website privacy policy corpus},
  author={Wilson, Shomir and Schaub, Florian and Dara, Aswarth Abhilash and Liu, Frederick and Cherivirala, Sushain and Leon, Pedro Giovanni and Andersen, Mads Schaarup and Zimmeck, Sebastian and Sathyendra, Kanthashree Mysore and Russell, N Cameron and others},
  booktitle={Proceedings of the 54th Annual Meeting of the Association for Computational Linguistics (Volume 1: Long Papers)},
  pages={1330--1340},
  year={2016}
}

@inproceedings {harkous2018polisis,
author = {Hamza Harkous and Kassem Fawaz and R{\'e}mi Lebret and Florian Schaub and Kang G. Shin and Karl Aberer},
title = {Polisis: Automated Analysis and Presentation of Privacy Policies Using Deep Learning},
booktitle = {27th USENIX Security Symposium (USENIX Security 18)},
year = {2018},
isbn = {978-1-939133-04-5},
address = {Baltimore, MD},
pages = {531--548},
url = {https://www.usenix.org/conference/usenixsecurity18/presentation/harkous},
publisher = {USENIX Association},
month = aug
}

@inproceedings{bannihatti2020finding,
  title={Finding a choice in a haystack: Automatic extraction of opt-out statements from privacy policy text},
  author={Bannihatti Kumar, Vinayshekhar and Iyengar, Roger and Nisal, Namita and Feng, Yuanyuan and Habib, Hana and Story, Peter and Cherivirala, Sushain and Hagan, Margaret and Cranor, Lorrie and Wilson, Shomir and others},
  booktitle={Proceedings of The Web Conference 2020},
  pages={1943--1954},
  year={2020}
}

@inproceedings{zhang2019bertscore,
  author       = {Tianyi Zhang and
                  Varsha Kishore and
                  Felix Wu and
                  Kilian Q. Weinberger and
                  Yoav Artzi},
  title        = {BERTScore: Evaluating Text Generation with {BERT}},
  booktitle    = {8th International Conference on Learning Representations, {ICLR} 2020,
                  Addis Ababa, Ethiopia, April 26-30, 2020},
  publisher    = {OpenReview.net},
  year         = {2020},
  url          = {https://openreview.net/forum?id=SkeHuCVFDr},
  bibsource    = {dblp computer science bibliography, https://dblp.org}
}

@inproceedings{NEURIPS2021_e4d2b6e6,
 author = {Yuan, Weizhe and Neubig, Graham and Liu, Pengfei},
 booktitle = {Advances in Neural Information Processing Systems},
 editor = {M. Ranzato and A. Beygelzimer and Y. Dauphin and P.S. Liang and J. Wortman Vaughan},
 pages = {27263--27277},
 publisher = {Curran Associates, Inc.},
 title = {BARTScore: Evaluating Generated Text as Text Generation},
 url = {https://proceedings.neurips.cc/paper/2021/file/e4d2b6e6fdeca3e60e0f1a62fee3d9dd-Paper.pdf},
 volume = {34},
 year = {2021}
}

@inproceedings{lin-2004-rouge,
    title = "{ROUGE}: A Package for Automatic Evaluation of Summaries",
    author = "Lin, Chin-Yew",
    booktitle = "Text Summarization Branches Out",
    month = jul,
    year = "2004",
    address = "Barcelona, Spain",
    publisher = "Association for Computational Linguistics",
    url = "https://aclanthology.org/W04-1013",
    pages = "74--81",
}

@inproceedings{grunewald2021tilt,
  title={TILT: A GDPR-aligned transparency information language and toolkit for practical privacy engineering},
  author={Gr{\"u}newald, Elias and Pallas, Frank},
  booktitle={Proceedings of the 2021 ACM Conference on Fairness, Accountability, and Transparency},
  pages={636--646},
  year={2021}
}

@article{DBLP:journals/corr/abs-2204-03556,
  author       = {Konrad Kollnig and
                  Anastasia Shuba and
                  Max Van Kleek and
                  Reuben Binns and
                  Nigel Shadbolt},
  title        = {Goodbye Tracking? Impact of iOS App Tracking Transparency and Privacy
                  Labels},
  journal      = {CoRR},
  volume       = {abs/2204.03556},
  year         = {2022},
  url          = {https://doi.org/10.48550/arXiv.2204.03556},
  doi          = {10.48550/ARXIV.2204.03556},
  eprinttype    = {arXiv},
  eprint       = {2204.03556},
  bibsource    = {dblp computer science bibliography, https://dblp.org}
}

@inproceedings{DBLP:conf/chi/LiRACH22,
  author       = {Tianshi Li and
                  Kayla Reiman and
                  Yuvraj Agarwal and
                  Lorrie Faith Cranor and
                  Jason I. Hong},
  editor       = {Simone D. J. Barbosa and
                  Cliff Lampe and
                  Caroline Appert and
                  David A. Shamma and
                  Steven Mark Drucker and
                  Julie R. Williamson and
                  Koji Yatani},
  title        = {Understanding Challenges for Developers to Create Accurate Privacy
                  Nutrition Labels},
  booktitle    = {{CHI} '22: {CHI} Conference on Human Factors in Computing Systems,
                  New Orleans, LA, USA, 29 April 2022 - 5 May 2022},
  pages        = {588:1--588:24},
  publisher    = {{ACM}},
  year         = {2022},
  url          = {https://doi.org/10.1145/3491102.3502012},
  doi          = {10.1145/3491102.3502012},
  bibsource    = {dblp computer science bibliography, https://dblp.org}
}

@article{gerl2018lpl,
  author       = {Armin Gerl and
                  Nadia Bennani and
                  Harald Kosch and
                  Lionel Brunie},
  title        = {{LPL}, Towards a GDPR-Compliant Privacy Language: Formal Definition
                  and Usage},
  journal      = {Trans. Large Scale Data Knowl. Centered Syst.},
  volume       = {37},
  pages        = {41--80},
  year         = {2018},
  url          = {https://doi.org/10.1007/978-3-662-57932-9\_2},
  doi          = {10.1007/978-3-662-57932-9\_2},
  bibsource    = {dblp computer science bibliography, https://dblp.org}
}

@article{gebru2021datasheets,
  title={Datasheets for datasets},
  author={Gebru, Timnit and Morgenstern, Jamie and Vecchione, Briana and Vaughan, Jennifer Wortman and Wallach, Hanna and Iii, Hal Daum{\'e} and Crawford, Kate},
  journal={Communications of the ACM},
  volume={64},
  number={12},
  pages={86--92},
  year={2021},
  publisher={ACM New York, NY, USA}
}

@article{cui2022pert,
  author       = {Yiming Cui and
                  Ziqing Yang and
                  Ting Liu},
  title        = {{PERT}: Pre-training {BERT} with Permuted Language Model},
  journal      = {CoRR},
  volume       = {abs/2203.06906},
  year         = {2022},
  url          = {https://doi.org/10.48550/arXiv.2203.06906},
  doi          = {10.48550/ARXIV.2203.06906},
  eprinttype   = {arXiv},
  eprint       = {2203.06906},
  bibsource    = {dblp computer science bibliography, https://dblp.org}
}

@inproceedings{sathyendra2017identifying,
  title={Identifying the provision of choices in privacy policy text},
  author={Sathyendra, Kanthashree Mysore and Wilson, Shomir and Schaub, Florian and Zimmeck, Sebastian and Sadeh, Norman},
  booktitle={Proceedings of the 2017 Conference on Empirical Methods in Natural Language Processing},
  pages={2774--2779},
  year={2017}
}

@article{liu2018towards,
  title={Towards automatic classification of privacy policy text},
  author={Liu, Frederick and Wilson, Shomir and Story, Peter and Zimmeck, Sebastian and Sadeh, Norman},
  journal={School of Computer Science Carnegie Mellon University},
  year={2018}
}

@inproceedings{nokhbeh2022privacycheck,
  title={PrivacyCheck v3: Empowering Users with Higher-Level Understanding of Privacy Policies},
  author={Nokhbeh Zaeem, Razieh and Ahbab, Ahmad and Bestor, Josh and Djadi, Hussam H and Kharel, Sunny and Lai, Victor and Wang, Nick and Barber, K Suzanne},
  booktitle={Proceedings of the Fifteenth ACM International Conference on Web Search and Data Mining},
  pages={1593--1596},
  year={2022}
}

@inproceedings{nokhbeh2020privacycheck,
  title={PrivacyCheck v2: A tool that recaps privacy policies for you},
  author={Nokhbeh Zaeem, Razieh and Anya, Safa and Issa, Alex and Nimergood, Jake and Rogers, Isabelle and Shah, Vinay and Srivastava, Ayush and Barber, K Suzanne},
  booktitle={Proceedings of the 29th ACM international conference on information \& knowledge management},
  pages={3441--3444},
  year={2020}
}

@article{ouyang2022training,
  title={Training language models to follow instructions with human feedback},
  author={Ouyang, Long and Wu, Jeffrey and Jiang, Xu and Almeida, Diogo and Wainwright, Carroll and Mishkin, Pamela and Zhang, Chong and Agarwal, Sandhini and Slama, Katarina and Ray, Alex and others},
  journal={Advances in Neural Information Processing Systems},
  volume={35},
  pages={27730--27744},
  year={2022}
}

@inproceedings{xue2020mt5,
  author       = {Linting Xue and
                  Noah Constant and
                  Adam Roberts and
                  Mihir Kale and
                  Rami Al{-}Rfou and
                  Aditya Siddhant and
                  Aditya Barua and
                  Colin Raffel},
  editor       = {Kristina Toutanova and
                  Anna Rumshisky and
                  Luke Zettlemoyer and
                  Dilek Hakkani{-}T{\"{u}}r and
                  Iz Beltagy and
                  Steven Bethard and
                  Ryan Cotterell and
                  Tanmoy Chakraborty and
                  Yichao Zhou},
  title        = {m{T}5: {A} Massively Multilingual Pre-trained Text-to-Text Transformer},
  booktitle    = {Proceedings of the 2021 Conference of the North American Chapter of
                  the Association for Computational Linguistics: Human Language Technologies,
                  {NAACL-HLT} 2021, Online, June 6-11, 2021},
  pages        = {483--498},
  publisher    = {Association for Computational Linguistics},
  year         = {2021},
  url          = {https://doi.org/10.18653/v1/2021.naacl-main.41},
  doi          = {10.18653/V1/2021.NAACL-MAIN.41},
  bibsource    = {dblp computer science bibliography, https://dblp.org}
}

@inproceedings{zhao2022fine,
  author       = {Kaifa Zhao and
                  Le Yu and
                  Shiyao Zhou and
                  Jing Li and
                  Xiapu Luo and
                  Aemon Yat Fei Chiu and
                  Yutong Liu},
  editor       = {Yoav Goldberg and
                  Zornitsa Kozareva and
                  Yue Zhang},
  title        = {A Fine-grained Chinese Software Privacy Policy Dataset for Sequence
                  Labeling and Regulation Compliant Identification},
  booktitle    = {Proceedings of the 2022 Conference on Empirical Methods in Natural
                  Language Processing, {EMNLP} 2022, Abu Dhabi, United Arab Emirates,
                  December 7-11, 2022},
  pages        = {10266--10277},
  publisher    = {Association for Computational Linguistics},
  year         = {2022},
  url          = {https://doi.org/10.18653/v1/2022.emnlp-main.700},
  doi          = {10.18653/V1/2022.EMNLP-MAIN.700},
  bibsource    = {dblp computer science bibliography, https://dblp.org}
}

@article{zimmeck2019maps,
  title={Maps: Scaling privacy compliance analysis to a million apps},
  author={Zimmeck, Sebastian and Story, Peter and Smullen, Daniel and Ravichander, Abhilasha and Wang, Ziqi and Reidenberg, Joel R and Russell, N Cameron and Sadeh, Norman},
  journal={Proc. Priv. Enhancing Tech.},
  volume={2019},
  pages={66},
  year={2019}
}

\appendix 
\label{sec:appendix}
\section{Key Information of APPSI-139}\label{Key-Information-of-APPSI-139}

\paragraph{Intended Users}
APPSI-139 is designed for researchers and practitioners in the fields of natural language processing, privacy policy analysis, and AI systems. It supports a wide range of applications, including multi-task learning, text summarization, and text classification. The dataset is also suitable for policymakers and legal scholars interested in examining how privacy practices are communicated in consumer-facing documents.

\paragraph{Hosting and Maintenance Plan}
All datasets in APPSI-139 are hosted and version-tracked via GitHub and are publicly available for direct download. Our core developing team is committed and has the resources to maintain and actively develop APPSI-139 for, at minimum, the next three years. We plan to grow APPSI-139 in several dimensions by including new learning datasets and leaderboards. We welcome external contributors. 

\paragraph{Computing resources} We use a server with an NVIDIA V100 GPU, Intel(R) Xeon(R) CPU with 128GB RAM for all empirical experiments in this manuscript.

\paragraph{Code availability} 

The source code, annotation guidelines, and dataset are provided in a public repository, which can be accessed via \url{https://github.com/EnlightenedAI/APPSI-139}.

\paragraph{Ethics Statement}
The development and dissemination of the APPSI-139 dataset adhere to stringent ethical standards to ensure the integrity of the data, and the responsible use of the information. This transparency ensures that users of the APPSI-139 dataset understand the origin of the data and the context in which it was collected. 

\paragraph{Transferability and Scalability}

The APPSI-139 dataset is designed with high transferability across different legal regimes and domains. Our annotation schema is hierarchical and fine-grained, abstracting common regulatory elements—such as data collection, sharing, and sensitive data categories—that are consistent across jurisdictions. For regulatory requirements in other regions, this schema is locally extendable or fine-tuneable without rebuilding the entire framework. In future work, our focus remains on exploring these localized extensions to support multilingual applications and cross-jurisdictional use.

\paragraph{Intended uses and dataset documentation }
The dataset is designed for automatic summarization and rewriting of privacy policies, aiming to help users make more informed and rational privacy decisions. By providing high-quality privacy policy summaries, users can more easily understand complex legal terms and privacy agreements, enabling them to make wiser choices when handling personal data and privacy rights.
The dataset supports research and implementation across various methods, including natural language classification and retrieval, text generation, in-context learning, and instruction fine-tuning. In addition to labeling and retrieving privacy policy content, it can also optimize LLMs through in-context learning and instruction fine-tuning, adjusting summary content based on user needs, thereby enhancing the model's performance in privacy policy applications.

Detailed documentation following the ``Datasheets for Datasets'' guidelines \citep{gebru2021datasheets} is provided in Section \ref{Datasheets for Datasets}.

\paragraph{Accessibility}
\begin{enumerate}
\item{Links to access the dataset and its metadata \url{https://github.com/EnlightenedAI/APPSI-139}.}

\item{The data is saved in a JSON format, where an example is shown in the README.md.}
\item{Research group will maintain this dataset on the official Github account.}
\item{CC BY 4.0. (\url{https://creativecommons.org/licenses/by/4.0/})}
\end{enumerate}

\section{Data Practice Category}\label{Data_Practice}
\textbf{\emph{Data Practice Category Information}} also known as \emph{Topic}, is used to describe the category of the sentence or term in privacy policies. It includes:

\begin{table*}
  \centering
  \begin{tabular}{l|p{10cm}}
\hline
\textbf{Category} & \textbf{Example Items} \\
\hline
Personal Property Information & Bank account, authentication information (password), deposit information (fund amounts, payment/receipt records), property information, credit records, credit reports, transaction and consumption records, transaction history, virtual currency, virtual transactions, game exchange codes, etc. \\
\hline

Personal Health Information & Illnesses, hospitalization records, medical prescriptions, laboratory reports, surgery/anesthesia records, nursing records, medication records, drug/food allergy information, reproductive information, past medical history, family medical history, infectious disease history, etc. \\
\hline
Biometric Information & Genetic data, fingerprints, voiceprints, palm prints, ear shapes, iris, facial recognition features, etc. \\
\hline
Personal Identity Information & ID card, military ID, passport, driver's license, work permit, social security card, residence permit, etc. \\
\hline
Other Information & Sexual orientation, marital history, religious beliefs, unpublished criminal records, communication records and content, contact lists, friend lists, group lists, travel history, web browsing history, accommodation information, precise location data, etc. \\
\hline
  \end{tabular}
  \caption{\label{Personal Sensitive Information}Examples of personal sensitive information.
  }
\end{table*}

\begin{itemize}[leftmargin=*]

\item[$\bullet$]{\textbf{First Party Collection}: The types of user information collected by the service provider, the purpose of collection, and whether providing such information is mandatory.}

\item[$\bullet$]{\textbf{Permission Acquisition}: How the service provider obtains application permissions from users, the purpose of these permissions, and whether they are mandatory.}
\item[$\bullet$]{\textbf{Third Party Sharing}: The purposes, types, and methods of sharing or disclosing user information to third parties.}
\item[$\bullet$]{\textbf{Usage}: How user data is used, including for data analysis, personalized recommendations, etc.}
\item[$\bullet$]{\textbf{Data Retention}: How user information is stored, including the duration and location of storage.}
\item[$\bullet$]{\textbf{Data Security}: How user information is protected and measures for handling data breaches.}
\item[$\bullet$]{\textbf{Edit/Control}: How users manage and handle their provided personal data.}
\item[$\bullet$]{\textbf{Specific Audiences}: Describes privacy practices specific to particular user groups, such as minors or residents of specific regions.}
\item[$\bullet$]{\textbf{Contact Information}: How to contact the service provider.}
\item[$\bullet$]{\textbf{Policy Change}: How users are notified about changes to the privacy policy.}
\item[$\bullet$]{\textbf{Cease Operation}: How user data is handled when the service ceases operations.}
\end{itemize}

\section{Personal Sensitive Information}\label{psi}
Personal sensitive information refers to data that, if disclosed, misused, or disclosed without authorization, may pose a threat to an individual’s safety and property. It can also lead to damage to personal reputation, physical and mental health, or result in discriminatory treatment. Generally, personal information related to children under 14 years old (inclusive) and any information directly linked to an individual's privacy is considered sensitive, as outlined in the Chinese national standard \textit{GB/T 35273-2020}. The following criteria can help determine whether information qualifies as personal sensitive information:

\begin{algorithm*}[ht]
\caption{TCSI-pp-V2 framework.}\label{alg:alg1}
\begin{algorithmic}
\STATE 
\STATE \textbf{Input:} {Privacy policy $P$; Specified $topics \in Topics$.}
\STATE \textbf{Output:} {Summarization $P_{ats}$.}
\STATE \textbf{Initialize:} {$P=\{p_1,...,p_n\} \gets Preprocessing (P) ; Filtered=list()$}
\STATE {\#Step 1: Five trained $experts$ carry out auto summarization.}
\STATE \textbf{For }$\forall p_j \in P:$
\STATE \hspace{0.5cm}{$f_j \gets F_f(F_e (p_j, \theta_e), \theta_f)$}

\STATE \hspace{0.5cm}{\textbf{If} $F_i (f_j, \theta_i)$ is $True$: } \hspace{0.2cm}{$topic = F_t(f_j,\theta_t)$ } 
\STATE \hspace{0.5cm}{\textbf{If} $\exists topic $ in $topics$:}

\STATE \hspace{1.1cm}{$Filtered \gets topic, r_j=F_r(f_j, \theta_r ) , s_j=F_s(f_j, \theta_s ), rewrite_j=F_{rewrite}(f_j, \theta_{rewrite})$ } 
cc
\STATE {\#Step 2: Examine into a more understandable Summarization.}
\STATE \textbf{For }{$\forall topic \in topics$:} {$P_{ats} \gets title\_topic$ \hspace{0.2cm} }
\STATE \hspace{0.5cm} {\#Write the title of $topic$ to $P_{ats}$.}
\STATE \hspace{0.5cm}\textbf{For }{$\forall rewrite_i \in topic$ in $Filtered $:}
\STATE \hspace{1.1cm}{\textbf{If} $r_i$ or $s_i$ is $True$:} \hspace{0.2cm}{$P_{ats} \gets rewrite_i$ with highlight} 
\STATE \hspace{1.1cm}{\textbf{Else}:}\hspace{0.2cm} {$P_{ats} \gets rewrite_i$)}
\STATE \textbf{Return} {$P_{ats}$}

\end{algorithmic}
\end{algorithm*}

\begin{table*}[ht]\centering

{
\begin{tabular}{ccccc}
\hline

{\color{black}} & \textbf{all samples} &	\textbf{training samples}   &  \textbf{validation samples}  & \textbf{testing samples} \\
\hline

Important &30877	&24710   &3088	  &3088    \\

Risk	  &15579    &12607	 &1576    &1576    \\

sensitive &15579    &12607	 &1576    &1576   \\

Topic    & 15579    &12607	 &1576    &1576\\

Rewritten &15677    &12543	 &1567    &1567    \\

\hline
\end{tabular}}
\caption{{\color{black} }The statistics of train-validation-test data.}
\label{tab:information extraction4}
\end{table*}%

\begin{enumerate}
    \item \textbf{Disclosure}: When personal information is disclosed, the individual and the organization or institution collecting or processing it lose control over its distribution, resulting in uncontrolled spreading and usage. In some cases, personal information, once leaked, may be used against the individual’s will or in conjunction with other data, posing a significant risk to the person’s rights. This type of information should be categorized as personal sensitive information. For example, if someone’s ID card is used to register a phone number or open a bank account without their permission, it should be considered personal sensitive information.
    
    \item \textbf{Illegal Provision}: Certain personal information becomes a significant risk to the individual’s rights when shared without consent, especially if it’s spread beyond the intended scope. Such information should be regarded as personal sensitive information. For instance, sexual orientation, banking details, and medical history related to infectious diseases should be considered sensitive if disclosed without consent.

    \item \textbf{Abuse}: Some personal information, when used beyond its authorized limits or for purposes other than originally intended, may pose a substantial risk to an individual's rights. This information should be classified as personal sensitive information. For example, if an insurance company uses a person’s health information to determine premiums for marketing purposes without obtaining the individual's consent, this is an abuse of personal sensitive information.
\end{enumerate}

We provides examples of personal sensitive information in Table \ref{Personal Sensitive Information}.

\section{TCSI-pp-V2 framework}\label{pseudocode}
The pseudocode of the TCSI-pp-V2 framework is shown in Algorithm~\ref{alg:alg1}.

\section{Data Splitting}\label{Splitting}
We split the APPSI-139 corpus into training data, validation data, and testing data at a ratio of 80:10:10. This approach effectively prevents potential data leakage and ensures the independence of the testing units. The details of each corpus are shown in Table~\ref{tab:information extraction4}. The source code, annotation guidelines, and dataset are available in a public repository at \url{https://github.com/EnlightenedAI/APPSI-139}.


\begin{table*}[t]\centering

\begin{tabular}{ccccccccccccc}
\hline
{\color{black}} \multirow{3}{*}{Topic}& \multicolumn{6}{c}{\textbf{TCSI-pp}} & \multicolumn{6}{c}{\textbf{TCSI-pp-v2}}\\

& \multicolumn{3}{c}{Xlnet} &\multicolumn{3}{c}{Electra} & \multicolumn{3}{c}{Xlnet2gpt}&\multicolumn{3}{c}{Electra2gpt}\\

{\color{black} } & P &	R  &F	 &P	 &R  &	F&P	 &R  &	F&P	 &R  &	F\\
\hline
First 
&81.5  &82.6  &82.1 
&88.2  &79.4  &83.5
&84.4  &80.3  &82.3       
&90.2  &71.6  &79.9
\\

Third  
&81.3  &78.1  &79.7       
&81.3  &78.1  &79.7 
&79.3  &82.4  &80.8
&81.0  &71.8  &76.1

\\

Usage  
&85.9  &78.9  &82.2  
&88.2  &80.6  &84.3
&88.7  &75.9  &81.9
&83.3  &80.6  &82.0
\\

Retention 
&81.6  &85.7  &87.9	      
&79.5  &79.5  &79.5
&80.0  &72.7  &76.2
&81.5  &75.0  &78.1
\\

Security 
&83.8  &76.0  &79.7       
&70.4  &74.2  &72.3
&73.2  &79.6  &76.3
&79.7  &67.7  &73.3
\\

Specific
&90.3  &85.7  &87.9	   
&89.5  &89.3  &89.4 
&91.2  &83.9  &87.4
&83.0  &91.6  &87.1
\\

Control 
&83.8  &76.0  &79.7	
&78.7  &78.7  &78.7
&89.5  &68.0  &77.3
&81.7  &77.3  &79.5
\\

Contact 
&87.1  &80.6  &83.7       
&91.5  &80.6  &85.7
&89.8  &79.1  &84.1
&94.6  &79.1  &86.2
\\

Change  
&75.8  &65.3  &70.1 
&75.7  &60.4  &67.2
&86.6  &49.3  &62.8
&73.2  &62.5  &67.4
\\
\hline
\end{tabular}
\caption{{\color{black} }Evaluation metrics for the Topics identification models.}
\label{tab:information extraction of topic}
\end{table*}%

\begin{table*}[ht]\centering

{
\begin{tabular}{cccc}
\hline
Characteristic & Attribute &Subtotal&Percentage\\
\hline
\multirow{2}{*}{Gender}  
& Male	  &27 &50.94\% \\
 &Female  &26 &49.06\%	    \\
\hline
 \multirow{5}{*}{Age}	 & < 18  &0& 0\%   \\

        &18-25   &38   &71.70\%    \\
	    &26-30	 &12   &22.64\%      \\
 	    &31-40	 &3	   &5.66\%      \\
        &>40     &0	   &0\%     \\
\hline
 \multirow{3}{*}{Education}	 &Associate \& below   &0	 &0\%         \\

        &Bachelor's &23   &43.39\%    \\

	    &Master's \& above	 &30	 &56.60\%      \\
\hline

\end{tabular}}
\caption{{\color{black} Basic statistical information of the survey participants.}}
\label{tab:interviewee}
\end{table*}

\section{Detailed Results for Topic Classification}\label{app:topic_extraction}

Table~\ref{tab:information extraction of topic} has provided a comprehensive evaluation of Precision, Recall, and F1 scores for multi-class classification on Topic, covering two models with the same encoder modules (XLNet and ELECTRA) under the TCSI-pp and TCSI-pp-V2 frameworks.

According to Table~\ref{tab:information extraction of topic}, XLNet and ELECTRA achieve comparable performance across topics under both the TCSI-pp and TCSI-pp-V2 frameworks. In terms of F1 scores, XLNet performs best overall, with scores above 0.79 for all topics except Policy Change. Across all metrics, models based on TCSI-pp and TCSI-pp-V2 show mixed results, with no significant differences, further indicating that TCSI-pp-V2's multi-task models match the performance of TCSI-pp while offering an efficiency advantage. Additionally, all models score relatively low on Policy Change, likely due to its small dataset proportion.

\section{Additional of Readability Survey}\label{ars}

\subsection{Basic statistical information of the survey participants}\label{interviewee}

Table~\ref{tab:interviewee} presents the basic statistical information of the survey participants in the privacy policy readability survey.

\begin{figure*}[t]\centering 
    \includegraphics[width=0.90\textwidth,trim={15cm 8cm 15cm 0cm}, clip]{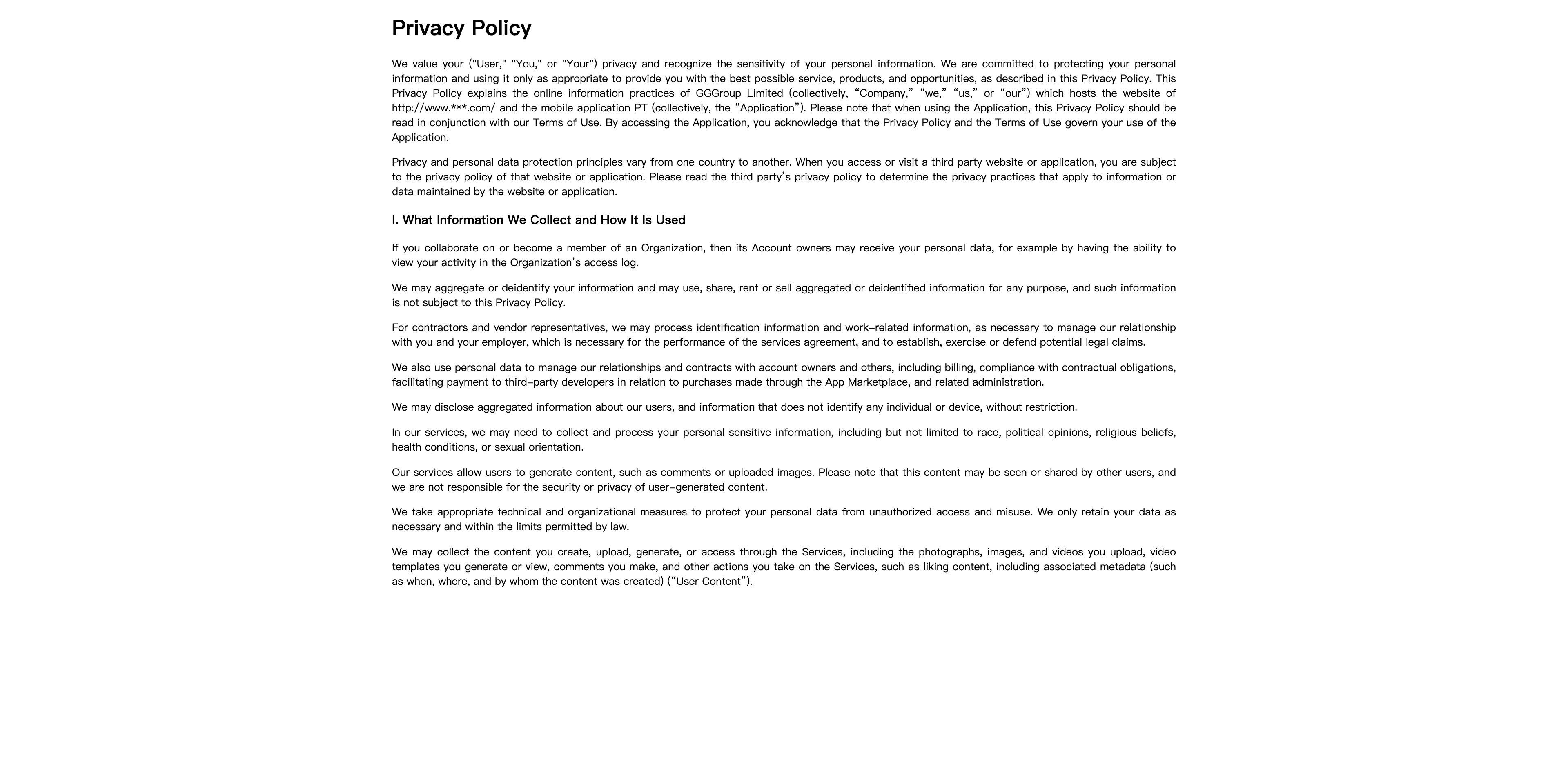}\vspace{-0em}
    \caption{Application Privacy
Policy.}
    \label{fig:pp}
    \vspace{-0em}
\end{figure*}

\begin{figure*}[t]\centering 
    \includegraphics[width=0.90\textwidth,trim={0cm 0cm 0cm 0cm}, clip]{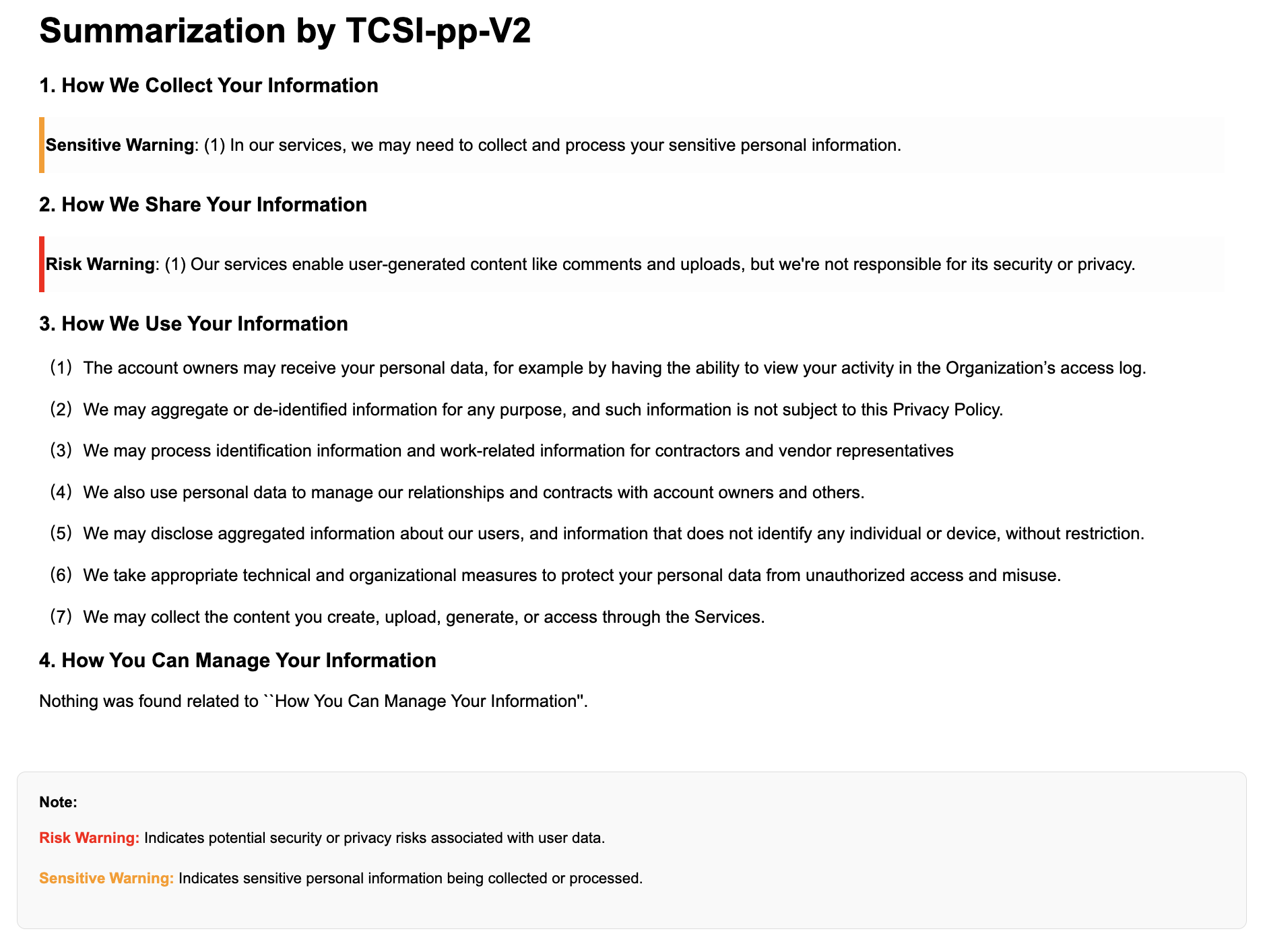}
    \caption{Summarization by TCSI-pp-V2.}
    \label{fig:our11}
    \vspace{-0em}
\end{figure*}

\begin{figure*}[t]\centering 
    \includegraphics[width=0.90\textwidth]{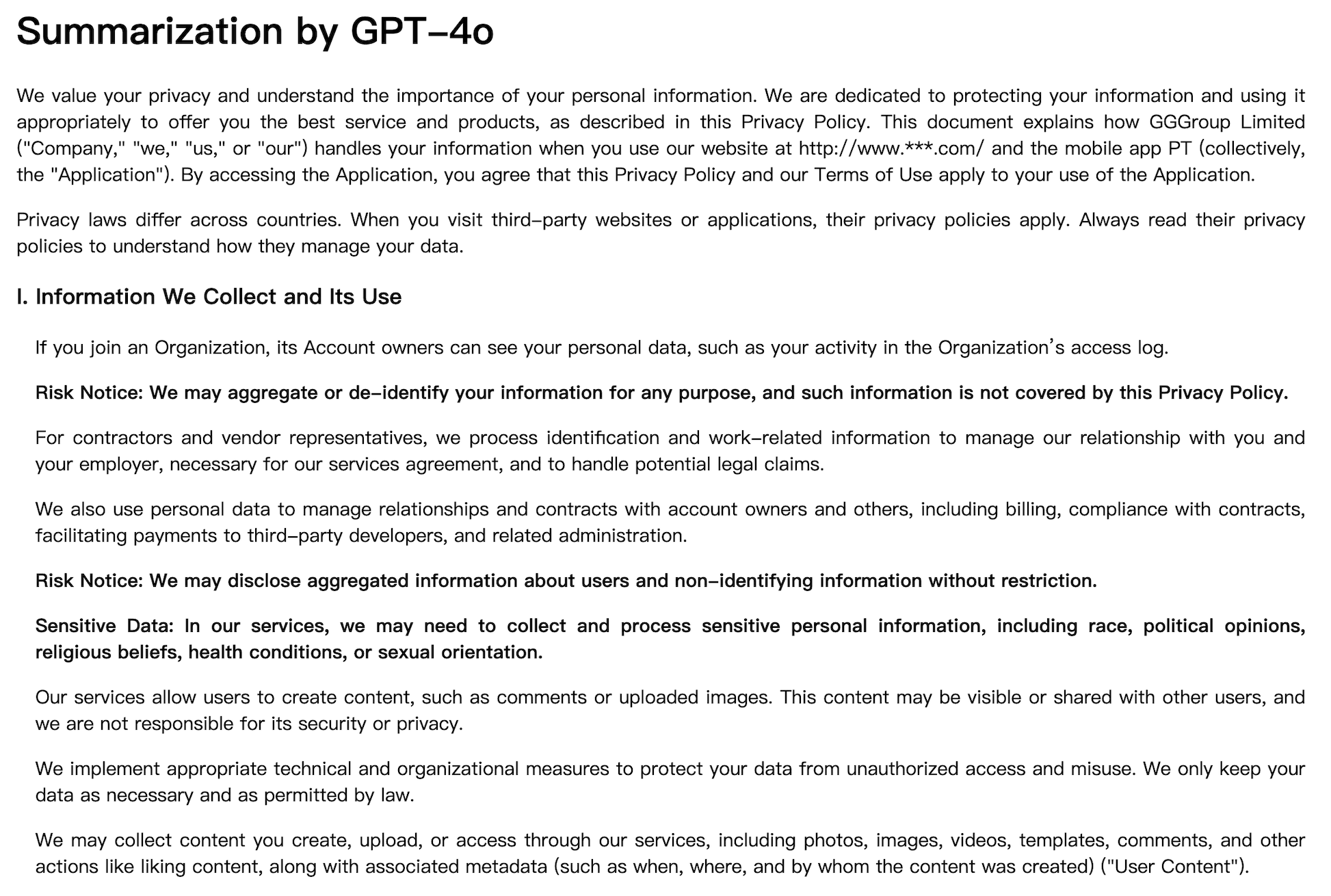}\vspace{0em}
    \caption{Summarization by GPT-4o.}
    \label{fig:gpt}
    \vspace{-0em}
\end{figure*}

\begin{figure*}[t]\centering 
    \includegraphics[width=0.90\textwidth]{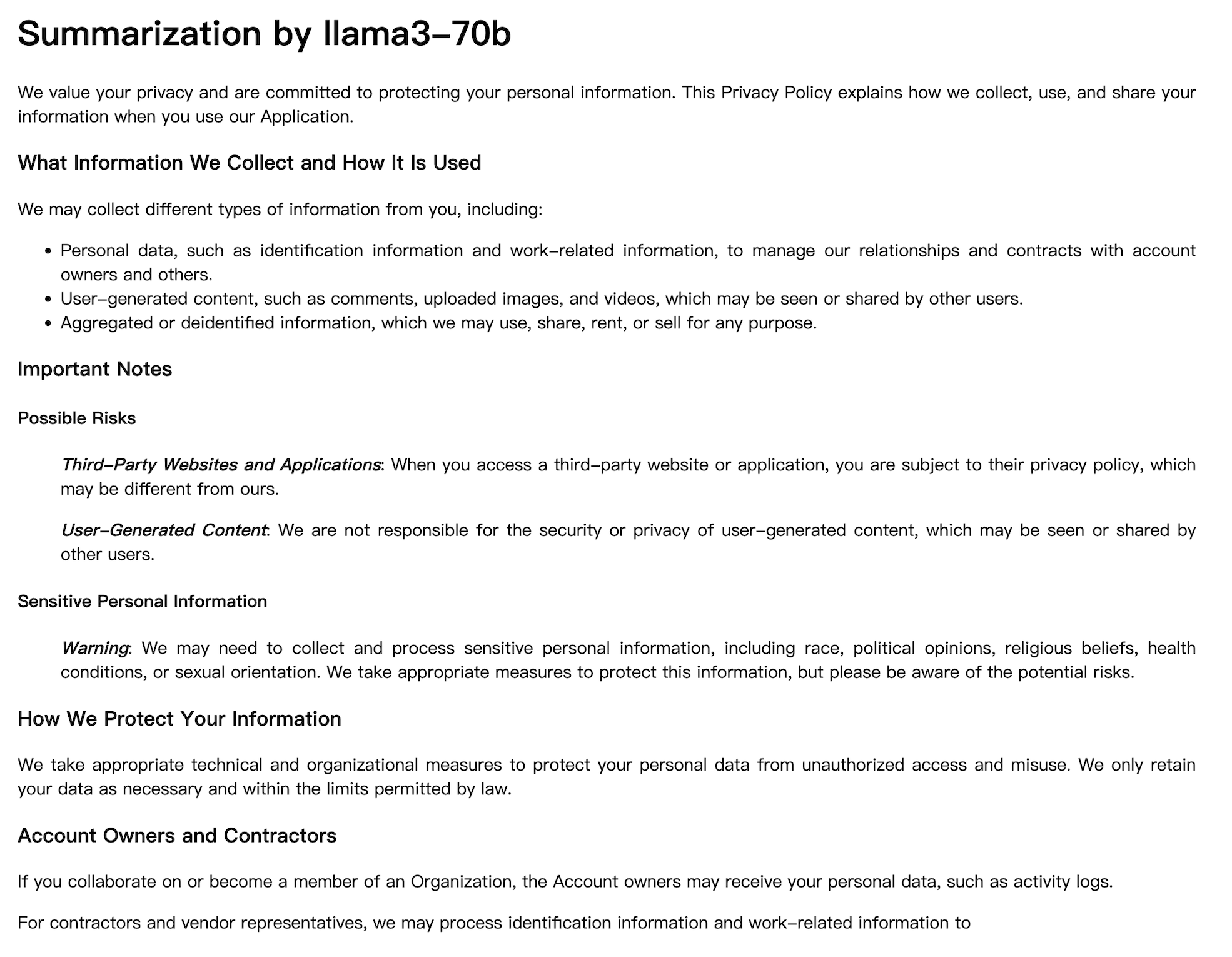}\vspace{-1em}
    \caption{Summarization by Llama3-70b.}
    \label{fig:llama}
    \vspace{-1em}
\end{figure*}

\begin{figure*}[t]\centering 
    \includegraphics[width=0.90\textwidth]{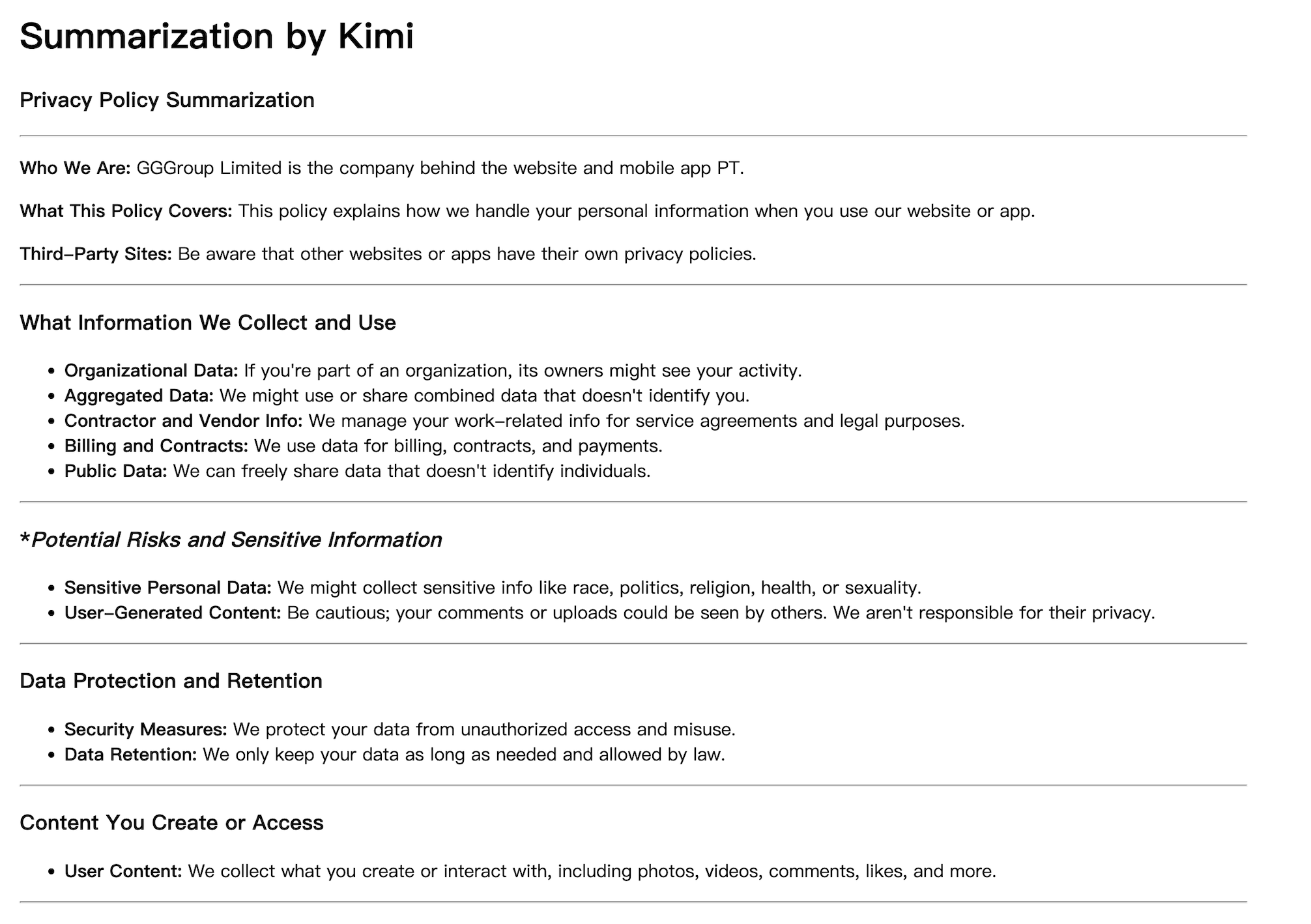}\vspace{0em}
    \caption{Summarization by Kimi (Moonshot V1).}
    \label{fig:kimi}
    \vspace{-0em}
\end{figure*}

\subsection{Examples of Summarization}\label{Examples}

To objectively assess the performance of TCSI-pp-V2, fine-tuned based on the mt5-small model, alongside large language models GPT-4o, Llama3-70b, and Kimi (Moonshot V1) in the task of privacy policy summary generation, we anonymized a privacy policy document and selected a portion of its content for testing, which is displayed in main paper Figure \ref{fig:pp}. We then presented the summary results of TCSI-pp-V2, GPT-4o, Llama3-70b, and Kimi (Moonshot V1), shown in Figures \ref{fig:our11}, \ref{fig:gpt}, \ref{fig:llama}, and \ref{fig:kimi}, respectively. The results indicate that TCSI-pp-V2 has a distinct advantage in generating structured summaries, capturing the essence of the original text with conciseness and precision. However, the summaries produced by other general language models appear more fragmented and lack coherence, as demonstrated by Llama3-70b and Kimi; or they exhibit shortcomings in identifying key information and effectively synthesizing it, such as GPT-4o, which almost verbatim reproduces the entire privacy policy document.

\subsection{Instruction Template and Evaluation Setup for LLMs.}

To ensure fairness and consistency, we have provided structured guidance to the large language models (LLMs) via the prompt template detailed in Algorithm \ref{alg:policy_summary}. This template includes comprehensive category definitions to establish clear task boundaries and explicit requirements for rewriting and formatting. In our comparative experiments, we utilize this standardized prompt with specific hyperparameter settings (e.g., temperature = 0.7 and top-p = 0.9) to ensure reproducible generation across models. These measures were implemented to minimize ambiguity and ensure that all models are evaluated under the same well-defined conditions.

\subsection{Evaluation Methodology}

To capture the overall preferences of human evaluators, we evaluate the summaries across four critical dimensions: \textit{comprehensibility} (how easy the summary is to understand), \textit{completeness} (whether the summary covers all key aspects without omissions), \textit{fidelity} (the accuracy and consistency of the summary with the original text, ensuring that no misunderstandings or distortions occur), and \textit{conciseness} (the ability to convey the essential points without unnecessary elaboration). For the readability survey, these dimensions are combined into a single composite score. While we acknowledge that this approach may obscure some fine-grained differences, the composite score effectively captures the trade-offs participants make across these dimensions and their holistic judgment of summary quality.

To ensure a rigorous and unbiased comparison, all summaries are normalized to maintain consistent formatting across different systems. During the survey, the summaries are presented to each evaluator in a randomized order to mitigate potential order biases. Furthermore, all materials are presented in a consistent monochrome format, aligned with the baseline methods. The colored highlights and nodes in Figure \ref{fig:our11} are provided solely as a visual aid for the readers and were absent from the actual human evaluation process.

A critical challenge in legal and privacy policy summarization is the trade-off between readability and factuality. While simplifying complex legal jargon enhances user comprehension, excessive abstraction may risk omitting nuanced legal obligations or distorting factual accuracy. Our framework addresses this by leveraging expert-annotated interpretation data, ensuring that the simplified output remains grounded in the original legal intent. Our human evaluation criteria—specifically fidelity and completeness—are designed to monitor this trade-off, ensuring that improved readability does not come at the expense of factual reliability.

\begin{algorithm*}[ht]
\caption{Privacy Policy Summarization Instruction Template.}\label{alg:policy_summary}
\begin{algorithmic}[1]
\STATE \textbf{Task Description:}
\STATE You are an expert in privacy policies. Please summarize the following privacy policy text sentence by sentence and classify each according to the predefined categories of data practices.

\STATE \textbf{Privacy Policy Categories:}
\STATE \hspace{0.5cm} \textit{First-Party Collection:} The types of user information collected by the service provider, the purposes of collection, and whether it is mandatory for users to provide this information.
\STATE \hspace{0.5cm} \textit{Permission Acquisition:} How the service provider obtains user permissions for the application, the purpose of these permissions, and whether these permissions are mandatory for users to grant.
\STATE \hspace{0.5cm} \textit{Third-Party Sharing/Disclosure:} The purposes, types, and methods of sharing or disclosing user information to third parties, and the compliance with regulations.
\STATE \hspace{0.5cm} \textit{----Other categories can be added as needed.----}

\STATE \textbf{Special Markings:}
\STATE \hspace{0.5cm} \textit{Risk:} Assigned to content that may raise compliance concerns, including potential violations of major data protection regulations (e.g., GDPR, CCPA) or vague descriptions of data handling practices.
\STATE \hspace{0.5cm} \textit{Sensitivity:} Covers data practices involving users’ sensitive personal information, such as biometric data, precise geolocation, financial account details, etc.

\STATE \textbf{Task Requirements:}
\STATE \hspace{0.5cm} \texttt{1.} Read and analyze the privacy policy text sentence by sentence or paragraph by paragraph, identifying and categorizing each part according to the relevant data practices.
\STATE\hspace{0.5cm} \texttt{2.} Summarize each data practice category using clear and simple language, ensuring that non-expert users can easily understand the core content.
\STATE \hspace{0.5cm} \texttt{3.} Use a multi-level bullet point structure to organize the summary results, making it easier to review and analyze. Use a concise, indented list structure.
\STATE \hspace{0.5cm} \texttt{4.} For clauses that involve “Risk” or “Sensitivity,” clearly mark them and briefly explain the potential issues or violations associated with them.

\STATE \textbf{Please summarize the following privacy policy based on the above instructions:}
\STATE \hspace{0.5cm} [----\textit{Insert privacy policy paragraph here}----]
\end{algorithmic}
\end{algorithm*}

\section{Datasheets for Datasets}\label{Datasheets for Datasets}
\subsection{Motivation}
\begin{enumerate}
\item[$\bullet$] \textbf{For what purpose was the dataset created?} (Was there a specific task in mind? Was there a specific gap that needed to be filled? Please provide a description.)

To address the common issues of ``lengthiness'' and ``incomprehensibility'' that the public often encounters when reading privacy policies, and to prevent information security risks caused by excessive authorization, we have carried out innovative work. Although existing English privacy policy corpora can alleviate the issue of ``lengthiness'' to some extent, they lack practical solutions for the ``incomprehensibility'' problem caused by professional jargon, technical terms, and complex sentence structures. Therefore, we have released the APPSI-139 corpus, which consists of 139 English application privacy policies and 30,877 sentences. In this corpus, we have annotated 11 types of data practices and three special markers (to identify the \emph{Importance}, \emph{Risk}, and \emph{Sensitivity} of privacy policies), and ultimately rewritten the important sentences through Text interpretation, i.e., reformulated them into a language form that is easier to understand. This is different from other English privacy policy summary datasets, as APPSI-139 provides rewritten content, contributing to the development of generative automatic summarization technology in the field of privacy policies. Based on APPSI-139, we have conducted benchmark tests on the performance of classic deep learning models in information extraction and text generation tasks.

\item[$\bullet$] \textbf{Who created this dataset (e.g., which team, research group) and on behalf of which entity (e.g., company, institution, organization)?}

DIVER, an interdisciplinary research group at Zhejiang University.

\end{enumerate}
\subsection{Composition}
\begin{enumerate}

\item[$\bullet$] \textbf{What do the instances that comprise the dataset represent (e.g., documents, photos, people, countries)} (Are there multiple types of instances (e.g., movies, users, and ratings; people and interactions between them; nodes and edges)? Please provide a description.)

The sentences in the privacy policy constitute instances of the APPSI-139, which include ``Original Sentence'', ``Data Practice Category'', ``Special Marking'', and the ``Rewritten Sentence''.

\item[$\bullet$] \textbf{How many instances are there in total} (of each type, if appropriate)?

The dataset contains 139 English privacy policies of mainstream applications, including 30877 preprocessed sentences.

\item[$\bullet$] \textbf{Does the dataset contain all possible instances or is it a sample (not necessarily random) of instances from a larger set?} (If the dataset is a sample, then what is the larger set? Is the sample representative of the larger set (e.g., geographic coverage)? If so, please describe how this representativeness was validated/verified. If it is not representative of the larger set, please describe why not (e.g., to cover a more diverse range of instances, because instances were withheld or unavailable).)

This dataset is a sample from a larger set. Details could be checked in Section \ref{subsec: dataset}.

\item[$\bullet$] \textbf{What data does each instance consist of? }(``Raw'' data (e.g., unprocessed text or images)or features? In either case, please provide a description.)

Each instance has ``Original Sentence'', ``Sentence ID'' and annotations, including ``Data Practices Category'', `` Special Marking'', and ``Rewritten Sentence''.

\item[$\bullet$] \textbf{Is there a label or target associated with each instance?} If so, please provide a description.

The privacy policy annotations are structured into three distinct components: ``Data Practices Category'', `` Special Marking'', and ``Rewritten Sentence''. For comprehensive details, refer to Section \ref{subsec: dataset} of the main paper.

\item[$\bullet$] \textbf{Is any information missing from individual instances?}(If so, please provide a description, explaining why this information is missing (e.g., because it was unavailable). This does not include intentionally removed information, but might include, e.g., redacted text.)

Sentences that lack special marking do not necessitate rewriting. As these sentences pertain to mundane declarative documents, summarization and rewriting are deemed unnecessary.

\item[$\bullet$] \textbf{Are relationships between individual instances made explicit (e.g., users' movie ratings, social network links)? }( If so, please describe how these relationships are made explicit.)

In affirmation, each instance is endowed with a distinctive identifier ``id'', which facilitates the correlation to its respective privacy policy. 

\item[$\bullet$] \textbf{Are there recommended data splits (e.g., training, development/validation, testing)? }(If so, please provide a description of these splits, explaining the rationale behind them.)

Yes. We did a train-validation-test split on the dataset, see Appendix \ref{Splitting}.

\item[$\bullet$] \textbf{Are there any errors, sources of noise, or redundancies in the dataset?} (If so, please provide a description.)

The creation of the APPSI-139 corpus involved annotations by legal experts, which may introduce bias based on their individual interpretations and perspectives. While efforts have been made to ensure consistency, such as strict annotation standards, variations in interpretations may impact the quality of the corpus.

\item[$\bullet$] \textbf{Is the dataset self-contained, or does it link to or otherwise rely on external resources (e.g., websites, tweets, other datasets)?}(If it links to or relies on external resources, a) are there guarantees that they will exist, and remain constant, over time; b) are there official archival versions of the complete dataset (i.e., including the external resources as they existed at the time the dataset was created); c) are there any restrictions (e.g., licenses, fees) associated with any of the external resources that might apply to a future user? Please provide descriptions of all external resources and any restrictions associated with them, as well as links or other access points, as appropriate.)

The APPSI-139 dataset under consideration is autonomous and does not necessitate or depend on external resources.

\item[$\bullet$] \textbf{Does the dataset contain data that might be considered confidential (e.g., data that is protected by legal privilege or by doctor-patient confidentiality, data that includes the content of individuals' non-public communications)?} (If so, please provide a description.)

No, all these privacy policies are publicly available.

\item[$\bullet$] \textbf{Does the dataset contain data that, if viewed directly, might be offensive, insulting, threatening, or might otherwise cause anxiety?} (If so, please describe why.)

No, the APPSI-139 does not include any data that could be considered offensive, insulting, threatening, or anxiety-provoking. This is because the dataset comprises privacy policies, and public documents to inform users and obtain their consent.

\item[$\bullet$] \textbf{Does the dataset relate to people?} (If not, you may skip the remaining questions in this section.)

No, the APPSI-139 does not pertain to individuals or personal data. It focuses solely on the content and structure of privacy policies.

\item[$\bullet$] \textbf{Does the dataset identify any subpopulations (e.g., by age, gender)?} If so, please describe how these subpopulations are identified and provide a description of their respective distributions within the dataset.

N/A.

\item[$\bullet$] \textbf{Is it possible to identify individuals (i.e., one or more natural persons), either directly or indirectly (i.e., in combination with other data) from the dataset?}(If so, please describe how how these subpopulations are identified and provide a description of their respective distributions within the dataset.)

N/A.

\item[$\bullet$] \textbf{Does the dataset contain data that might be considered sensitive in any way (e.g., data that reveals racial or ethnic origins, sexual orientations, religious beliefs, political opinions or union memberships, or locations; financial or health data; biometric or genetic data; forms of government identification, such as social security numbers; criminal history)?} (If so, please provide a description.)

N/A.

\item[$\bullet$] \textbf{Any other comments?}

N/A.

\end{enumerate}

\subsection{Collection Process}
\begin{enumerate}
\item[$\bullet$] \textbf{How was the data associated with each instance acquired?} (Was the data directly observable (e.g., raw text, movie ratings), reported by subjects (e.g., survey responses), or indirectly inferred/derived from other data (e.g., part-of-speech tags, model-based guesses for age or language)? If data was reported by subjects or indirectly inferred/derived from other data, was the data validated/verified? If so, please describe how.)

Please check Section \ref{subsec: dataset} in the main paper.

\item[$\bullet$] \textbf{What mechanisms or procedures were used to collect the data (e.g., hardware apparatus or sensor, manual human curation, software program, software API)?} (How were these mechanisms or procedures validated?)

We have developed a straightforward web crawler application to systematically retrieve public privacy policies from the Google Play Store and the Apple App Store. This program ensures a streamlined and efficient process for accessing these essential documents.

\item[$\bullet$] \textbf{If the dataset is a sample from a larger set, what was the sampling strategy (e.g., deterministic, probabilistic with specific sampling probabilities)?}

We utilized web crawling to collect English privacy policies from the top 100 apps in the U.S. on both the Google Play Store and Apple App Store. After deduplication and content verification, we obtained 139 complete application privacy policies, which were then segmented into sentence-level samples using regular expressions.

\item[$\bullet$] \textbf{Who was involved in the data collection process (e.g., students, crowdworkers, contractors) and how were they compensated (e.g., how much were crowdworkers paid)?}

All annotators received appropriate compensation according to standard practice, significantly above the local average wage.

\item[$\bullet$] \textbf{Over what timeframe was the data collected? (Does this timeframe match the creation timeframe of the data associated with the instances (e.g., recent crawl of old news articles)?} If not, please describe the timeframe in which the data associated with the instances was created.)

We collect privacy policy updates until October 2023.

\item[$\bullet$] \textbf{Were any ethical review processes conducted (e.g., by an institutional review board)?} (If so, please provide a description of these review processes, including the outcomes, as well as a link or other access point to any supporting documentation.)

Yes. We have conducted an internal ethical review process by the ethical team.

\item[$\bullet$] \textbf{Does the dataset relate to people?} (If not, you may skip the remaining questions in this section.)

No, the APPSI-139 does not pertain to individuals or personal data.

\item[$\bullet$] \textbf{Did you collect the data from the individuals in question directly, or obtain it via third parties or other sources (e.g., websites)?}Were the individuals in question notified about the data collection? If so, please describe(or show with screenshots or other information) how notice was provided, and provide a link or other access point to, or otherwise reproduce, the exact language of the notification itself.

N/A.

\item[$\bullet$] \textbf{Were the individuals in question notified about the data collection?}
(If so, please describe (or show with screenshots or other information) how notice was provided, and provide a link or other access point to, or otherwise reproduce, the exact language of the notification itself.)

N/A.

\item[$\bullet$] \textbf{Did the individuals in question consent to the collection and use of their data?} (If so, please describe (or show with screenshots or other information) how consent was requested and provided, and provide a link or other access point to, or otherwise reproduce, the exact language to which the individuals consented.)

N/A.

\item[$\bullet$] \textbf{If consent was obtained, were the consenting individuals provided with a mechanism to revoke their consent in the future or for certain uses?} (If so, please provide a description, as well as a link or other access point to the mechanism (if appropriate).)

N/A.

\item[$\bullet$] \textbf{Has an analysis of the potential impact of the dataset and its use on data subjects (e.g., a data protection impact analysis) been conducted?} (If so, please provide a description of this analysis, including the outcomes, as well as a link or other access point to any supporting documentation.)

N/A.

\item[$\bullet$] \textbf{Any other comments?}

N/A.

\end{enumerate}

\subsection{Preprocessing/cleaning/labeling}

\begin{enumerate}

\item[$\bullet$] \textbf{Was any preprocessing/cleaning/labeling of the data done (e.g., discretization or bucketing, tokenization, part-of-speech tagging, SIFT feature extraction, removal of instances, processing of missing values)?} (If so, please provide a description. If not, you may skip the remainder of the questions in this section.)

We removed the HTML tags because they are meaningless for privacy policy research.

\item[$\bullet$] \textbf{Was the ``raw'' data saved in addition to the preprocessed/cleaned/labeled data (e.g., to support unanticipated future uses)?} (If so, please provide a link or other access point to the ``raw'' data.)

Yes. We have archived the ``raw'' data on GitHub, ensuring its accessibility and preservation for future reference and analysis.

\item[$\bullet$] \textbf{Is the software used to preprocess/clean/label the instances available?} (If so, please provide a link or other access point.)

The entire annotation process was conducted using the Doccano~\citep{doccano} tool, which is accessible via the following link: \url{https://github.com/doccano/doccano}. Visual demonstrations of the annotation process are depicted in Figures \ref{fig:doccano} and \ref{fig:doccano1}. To foster transparency and reproducibility, we provide the source code, annotation guidelines, and dataset in a public repository, which can be accessed via \url{https://github.com/EnlightenedAI/APPSI-139}.

\begin{figure*}[t]
    \centerline{
    \includegraphics[width=0.9 \textwidth]{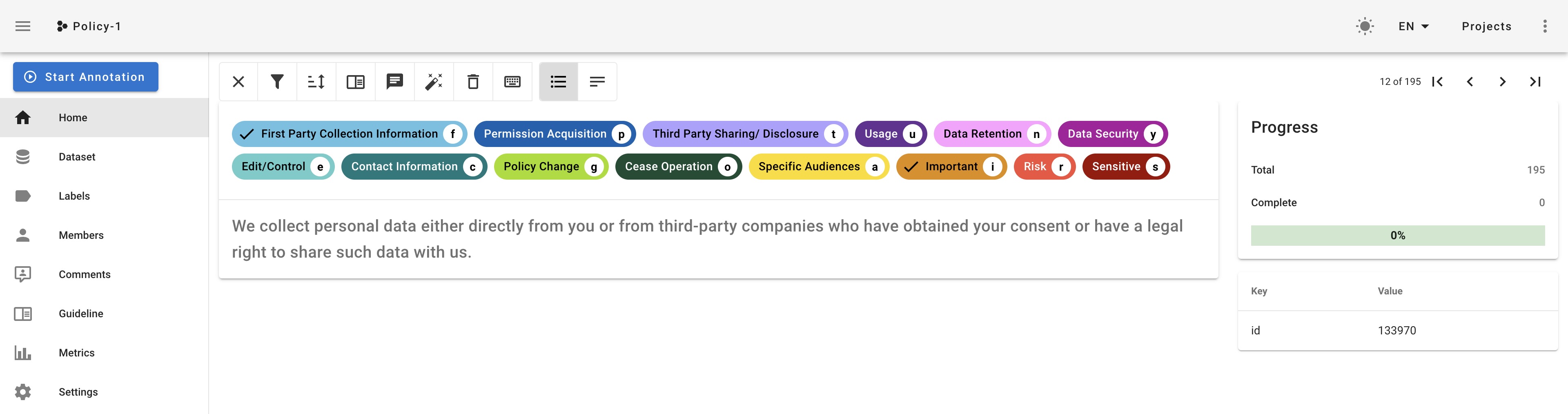}}
    \caption{Annotation of Privacy Policy in Doccano}
    \label{fig:doccano}
\end{figure*}

\begin{figure*}[t]\centering
    \includegraphics[width=0.9 \textwidth]{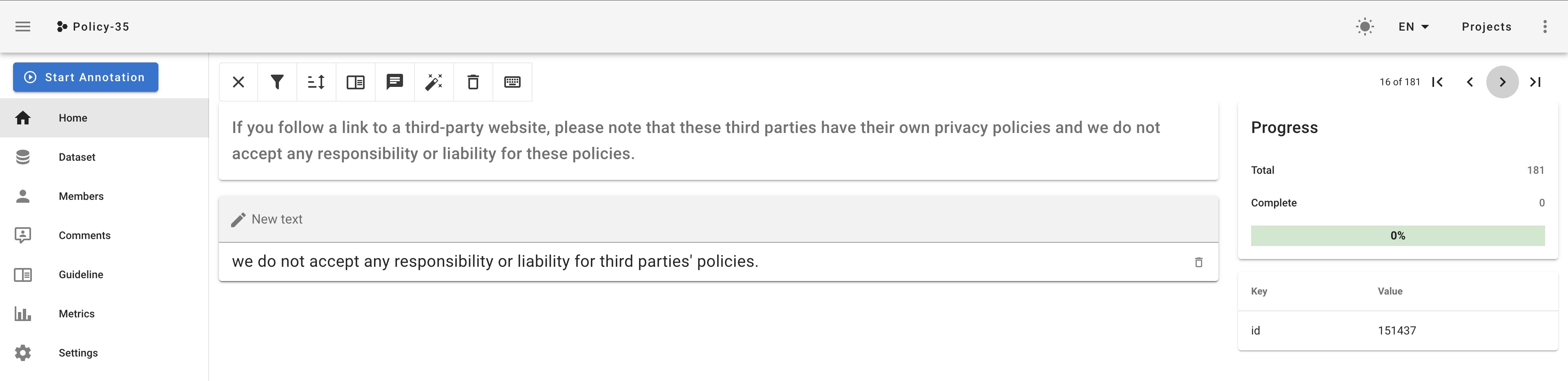}
    \caption{Rewritten of Privacy Policy in Doccano}
    \label{fig:doccano1}
\end{figure*}

\item[$\bullet$] \textbf{Any other comments?}

None.
\end{enumerate}

\subsection{Uses}
\begin{enumerate}
\item[$\bullet$] \textbf{Has the dataset been used for any tasks already?} (If so, please provide a description.)

The APPSI-139 dataset represents a novel resource for English application privacy policy summarization and interpretation. We have conducted extensive experiments to benchmark the performance of both classical machine learning algorithms and cutting-edge deep learning models across tasks such as automatic text summarization, text generation, and text classification. The comprehensive findings of these experiments are elaborated in Section \ref{Experiments} of our primary research document.

\item[$\bullet$] \textbf{Is there a repository that links to any or all papers or systems that use the dataset?} (If so, please provide a link or other access point.)

We provide the source code, annotation guidelines, and dataset in a public repository, which can be accessed via \url{https://github.com/EnlightenedAI/APPSI-139}.

\item[$\bullet$] \textbf{What (other) tasks could the dataset be used for?}

This dataset supports research and implementation across a variety of methods, including natural language classification and retrieval, text generation, in-context learning (including zero-shot and few-shot), and instruction fine-tuning. Beyond labeling and retrieving privacy policy content, the dataset can also optimize large language models through in-context learning and instruction fine-tuning, adjusting summary content based on user needs and thus improving the model's performance in privacy policy applications. The source code and dataset are provided in a public repository, which can be accessed via \url{https://github.com/EnlightenedAI/APPSI-139}.

\item[$\bullet$] \textbf{Is there anything about the composition of the dataset or the way it was collected and preprocessed/cleaned/labeled that might impact future uses?} (For example, is there anything that a future user might need to know to avoid uses that could result in unfair treatment of individuals or groups (e.g., stereotyping, quality of service issues) or other undesirable harms (e.g., financial harms, legal risks) If so, please provide a description. Is there anything a future user could do to mitigate these undesirable harms?)

No.

\item[$\bullet$] \textbf{Are there tasks for which the dataset should not be used?} (If so, please provide a description.)

No.

\item[$\bullet$] \textbf{Any other comments?}

N/A.

\end{enumerate}

\subsection{Distribution}
\begin{enumerate}
\item[$\bullet$] \textbf{Will the dataset be distributed to third parties outside of the entity (e.g., company, institution, organization) on behalf of which the dataset was created?} (If so, please provide a description.)

No.

\item[$\bullet$] \textbf{How will the dataset will be distributed (e.g., tarball on website, API, GitHub)?} (Does the dataset have a digital object identifier (DOI)?)

It is released on Github at 
\url{https://github.com/EnlightenedAI/APPSI-139}.

\item[$\bullet$] \textbf{When will the dataset be distributed?}

Before the conference.

\item[$\bullet$] \textbf{Will the dataset be distributed under a copyright or other intellectual property (IP) license, and/or under applicable terms of use (ToU)?} (If so, please describe this license and/or ToU, and provide a link or other access point to, or otherwise reproduce, any relevant licensing terms or ToU, as well as any fees associated with these restrictions.)

CC BY 4.0. (\url{https://creativecommons.org/licenses/by/4.0/})

\item[$\bullet$] \textbf{Have any third parties imposed IP-based or other restrictions on the data associated with the instances?} (If so, please describe these restrictions, and provide a link or other access point to, or otherwise reproduce, any relevant licensing terms, as well as any fees associated with these restrictions.)

No.

\item[$\bullet$] \textbf{Do any export controls or other regulatory restrictions apply to the dataset or to individual instances?} (If so, please describe these restrictions, and provide a link or other access point to, or otherwise reproduce, any supporting documentation.)

No.

\item[$\bullet$] \textbf{Any other comments?}

N/A.

\end{enumerate}

\subsection{Maintenance}
\begin{enumerate}
\item[$\bullet$] \textbf{Who is supporting/hosting/maintaining the dataset?}

DIVER, an interdisciplinary research group at Zhejiang University.

\item[$\bullet$] \textbf{How can the owner/curator/manager of the dataset be contacted (e.g., email address)?}

E-mail addresses are at the top of this document.

\item[$\bullet$] \textbf{Is there an erratum? }(If so, please provide a link or other access point.)

No.

\item[$\bullet$] \textbf{Will the dataset be updated (e.g., to correct labeling errors, add new instances, delete instances')?} (If so, please describe how often, by whom, and how updates will be communicated to users (e.g., mailing list, GitHub)?)

No. If we plan to update the dataset in the future, we will elaborate on the reason on our GitHub repository.

\item[$\bullet$] \textbf{If the dataset relates to people, are there applicable limits on the retention of the data associated with the instances (e.g., were individuals in question told that their data would be retained for a fixed period of time and then deleted)?} (If so, please describe these limits and explain how they will be enforced.)

No.

\item[$\bullet$] \textbf{Will older versions of the dataset continue to be supported/hosted/maintained?} (If so, please describe how. If not, please describe how its obsolescence will be communicated to users.)

Yes. If we plan to update the data, we will maintain the old version and then release the follow-up version, for example, APPSI-139-V2.0.

\item[$\bullet$] \textbf{If others want to extend/augment/build on/contribute to the dataset, is there a mechanism for them to do so?} (If so, please provide a description. Will these contributions be validated/verified? If so, please describe how. If not, why not? Is there a process for communicating/distributing these contributions to other users? If so, please provide a description.)

Yes. For data annotation, researchers could carefully check our annotation guidelines in GitHub to ensure consistency. And if others want to contribute to the dataset, they could submit a pull request or contact us via email.

\item[$\bullet$] \textbf{Any other comments?}

N/A.

\end{enumerate}
\end{document}